\documentclass[preprint,12pt,number]{elsarticle}

\biboptions{sort&compress}
\usepackage[a4paper,
            bindingoffset=0.2in,
            left=0.5in,
            right=0.5in,
            top=0.7in,
            bottom=0.7in,
            footskip=.25in]{geometry}

\usepackage{array}
\usepackage{booktabs}
\usepackage{multirow}
\usepackage{siunitx}
\usepackage{verbatim}
\usepackage{enumitem}
\usepackage[hidelinks]{hyperref}
\usepackage{adjustbox}
\usepackage{color,soul}
\usepackage{caption}
\usepackage{subcaption}
\usepackage{array}
\usepackage[dvipsnames]{xcolor}
\usepackage[table]{xcolor}

\usepackage[utf8]{inputenc}
\usepackage{mathtools, nccmath}

\usepackage{amsmath, amsfonts}
\usepackage{amssymb}
\usepackage[mathlines]{lineno}
\usepackage{lineno}
\usepackage{makecell}

\usepackage{textcomp}
\usepackage{graphicx}
\usepackage{algorithm} 
\usepackage{algpseudocode}
\usepackage{algorithmicx}
\usepackage[figurename=Fig.]{caption}

\usepackage{adjustbox}

\usepackage{wrapfig}
\DeclareCaptionLabelFormat{custom}{\textbf{#1 #2}}
\DeclareCaptionLabelSeparator{custom}{. }
\DeclareCaptionFormat{custom}{#1#2 #3}
\makeatletter
\def\ps@pprintTitle{%
  \let\@oddhead\@empty%
  \let\@evenhead\@empty%
  \def\@oddfoot{}%
  \let\@evenfoot\@oddfoot}

\begin{document}

\begin{frontmatter}

\title{Physics-Guided Fully Convolutional Spatiotemporal Learning Toward Digital-Twin-Enabled Microstructure Evolution Prediction}

\author{{Michael Trimboli}}

\author{{Wenxi Liu}}

\author{Xianqi Li\corref{cor1}}

\cortext[cor1]{Corresponding author}

\affiliation{organization={Department of Mathematics and Systems Engineering},
            addressline={Florida Institute of Technology}, 
            city={Melbourne},
            state={FL},
            postcode={32901}, 
            country={USA}}
            
\begin{abstract}
Understanding and predicting microstructure evolution is central to materials design, yet purely data-driven spatiotemporal learning models often suffer from limited physical consistency and degraded long-term prediction accuracy. In this work, we introduce a physics-guided fully convolutional spatiotemporal learning framework for microstructure evolution prediction. Unlike prior self-supervised approaches, the proposed method explicitly incorporates governing physical equations into the training objective, thereby encouraging the learned dynamics to remain consistent with known thermodynamic and kinetic laws. This physics-guided formulation improves predictive accuracy, long-horizon stability, and robustness across spatial resolutions and temporal prediction settings. Extensive experiments for spinodal decomposition demonstrate that incorporating physics-guided residual regularization leads to more faithful reproduction of microstructural morphology, statistics, and evolution trends compared with purely data-driven baselines. The proposed framework preserves the scalability and computational efficiency of fully convolutional architectures while bridging the gap between high-fidelity physics-based simulations and data-driven surrogate modeling, offering a reliable and efficient surrogate-modeling step toward digital-twin-enabled microstructure evolution prediction.

\end{abstract}


\end{frontmatter}


\section{Introduction}

Microstructure evolution is a central problem in computational materials science because the spatial organization of phases, grains, interfaces, and domains strongly influences the resulting properties and performance of materials. During processing and service, microstructures evolve through thermodynamically driven and kinetically constrained mechanisms, leading to changes in morphology, connectivity, interface geometry, and characteristic length scales \cite{chen2002phase,wang2005property,tourret2022phase}. Accurate prediction of these changes is essential for understanding processing--microstructure--property relationships and for enabling accelerated materials design. 
Among the many forms of microstructure evolution, spinodal decomposition is a canonical phase-separation process in which a homogeneous binary mixture spontaneously separates into compositionally distinct domains after being quenched into a thermodynamically unstable regime \cite{cahn1961spinodal,oono1988study}. The evolution involves the amplification of composition fluctuations, formation of diffuse interfaces, and subsequent coarsening of phase-separated domains. Because the resulting morphology directly affects macroscopic material behavior, spinodal decomposition has served as an important model problem for studying nonlinear pattern formation, diffusion-driven coarsening, and phase-field microstructure evolution \cite{cahn1961spinodal,wise2005surface,cates2018theories,chen2002phase}.

Phase-field modeling provides a physically rigorous framework for simulating such processes \cite{chen2002phase,tourret2022phase,wang2024systematic}. In particular, the Cahn--Hilliard (CH) equation describes the evolution of a conserved concentration field driven by chemical free-energy minimization and interfacial energy effects \cite{cahn1961spinodal}. This formulation naturally captures diffuse interfaces, topological changes, and long-time coarsening behavior without explicitly tracking phase boundaries. However, high-fidelity phase-field simulation requires repeated numerical solution of nonlinear, fourth-order partial differential equations over fine spatial grids and long temporal horizons. This computational cost becomes a major limitation when simulations must be repeated many times for parameter studies, uncertainty quantification, inverse design, process optimization, or digital-twin-enabled materials design.

A large body of work has sought to reduce this computational burden. One direction improves the efficiency of physics-based simulation itself through high-performance computing, adaptive numerical methods, efficient time integration, and parallel implementations \cite{hunter2011large,vondrous2014parallel,shi2017accelerating,du2020phase}. These approaches preserve the governing equations and therefore retain strong physical fidelity. Nevertheless, they still require solving the underlying nonlinear PDEs over the full computational domain, and their cost can remain substantial for high-resolution, long-time, or many-query simulation settings. Thus, while advanced solvers are indispensable for generating reliable reference data, they may be too expensive for repeated forecasting and rapid decision-making workflows. 
A second direction develops data-driven surrogate models that learn microstructure evolution directly from simulation data \cite{zhang2020machine,DOZapiain,yang2021self,hu2022accelerating,oommen2022learning,farizhandi2023spatiotemporal,ahmad2023accelerating}. These methods aim to replace repeated numerical time integration with neural network inference. Recurrent spatiotemporal architectures, including ConvLSTM-type models \cite{shi2015convolutional}, PredRNN++ \cite{wang2018predrnn++}, and E3D-LSTM \cite{wang2018eidetic}, have been used to predict phase-field microstructure evolution and have demonstrated encouraging performance on benchmark problems such as grain growth and spinodal decomposition \cite{yang2021self,hu2022accelerating,farizhandi2023spatiotemporal,ahmad2023accelerating}. By propagating hidden states over time, recurrent models naturally represent sequential dynamics. However, this same sequential structure limits temporal parallelization, increases inference latency for long prediction horizons, and may introduce accumulated hidden-state drift during recursive rollout. These limitations become especially important when surrogate models are intended for high-throughput forecasting, parameter sweeps, or integration into digital-twin pipelines.

Other surrogate-modeling paradigms have also been explored for PDE-driven materials dynamics. Autoencoder-based latent-space models compress high-dimensional microstructure fields into lower-dimensional representations and learn the evolution in the latent space \cite{hu2022accelerating,ahmad2023accelerating}. Such models can improve computational efficiency, but the compression step may remove fine interfacial information that is important for morphology-sensitive evolution. Neural operator methods, including Fourier Neural Operators, learn mappings between function spaces and can provide strong generalization for certain classes of parameterized PDEs \cite{oommen2022learning,li2020fourier}. However, global spectral representations may be less naturally aligned with microstructure evolution dominated by localized interface motion, droplet coarsening, and heterogeneous morphology changes. Graph neural networks offer another perspective by representing grains, particles, or microstructural features as nodes and edges \cite{qin2024graingnn}. While this representation can encode topological relationships, it often requires feature extraction, graph construction, and dynamic graph updates as the microstructure evolves, which can complicate direct dense-field prediction.

Physics-informed and physics-guided neural approaches provide a more principled route for incorporating governing equations into learning. Physics-informed neural networks (PINNs) enforce PDE residuals through the loss function and have become an influential framework for scientific machine learning \cite{raissi2019physics}. However, directly applying PINN-type formulations to high-dimensional phase-field microstructure evolution remains challenging. The CH equation is nonlinear, fourth order, and diffusion coupled; evaluating strong-form residuals over dense image sequences can be computationally expensive, especially when the goal is not to solve a single initial-boundary-value problem, but to learn a reusable forecasting model over many trajectories. Moreover, standard PINNs are typically designed for continuous-coordinate solution approximation, whereas microstructure surrogate modeling often requires fast multi-frame prediction on dense image grids. Thus, although physics-informed learning offers a natural path toward physical consistency, its integration with efficient, high-throughput spatiotemporal forecasting remains nontrivial.

More recently, fully convolutional spatiotemporal models have provided an efficient alternative to recurrent forecasting. Architectures such as SimVP \cite{gao2022simvp} and SimVPv2 \cite{tan2022simvpv2} replace sequential hidden-state propagation with convolutional spatiotemporal translation, allowing future frames to be predicted through a direct and highly parallel forward pass \cite{tan2022simvpv2}. Recent work in \cite{trimboli2026fully} showed that fully convolutional, non-recurrent spatiotemporal models can accurately predict grain growth and spinodal decomposition, generalize from low-resolution training data to higher-resolution simulations, and substantially reduce inference cost compared with recurrent baselines. However, this prior framework remained purely data-driven: the model was trained primarily to minimize image-level discrepancy between predicted and reference microstructure fields. While this strategy can produce reasonably accurate forecasts, it does not explicitly constrain the learned dynamics to satisfy the governing phase-field equation. 
This limitation is particularly important for spinodal decomposition, which is governed by strict physical structure, including mass conservation, chemical-potential-driven diffusion, interfacial energy, and nonlinear coarsening dynamics. A surrogate model may achieve low root-mean-square error (RMSE) or high structural similarity index measure (SSIM) while still producing trajectories that gradually deviate from physically admissible evolution. Such deviations may be subtle at the pixel level but become significant in long-horizon rollout, reduced-context prediction, particle statistics, curvature distributions, and coarsening trends. Therefore, the key challenge is no longer only whether a neural network can forecast microstructure images efficiently, but whether it can do so while remaining consistent with the thermodynamic and kinetic laws governing the evolution. 
This observation motivates the present work. Rather than treating microstructure forecasting as a purely data-driven prediction problem, we develop a \emph{physics-guided fully convolutional spatiotemporal learning framework} for spinodal decomposition prediction. The proposed approach retains the computational advantages of a lightweight fully convolutional backbone, while incorporating the governing CH equation into the training objective through a physics-guided regularization term. In this way, the model is encouraged not only to match the reference simulation data, but also to produce future microstructure sequences that remain more consistent with the underlying phase-field dynamics. Importantly, the physics guidance is introduced during training only; at inference time, prediction is still performed by a direct fully convolutional forward pass.

The main contribution of this study is therefore not a new phase-field solver, nor simply another deep learning benchmark for spinodal decomposition. Instead, we address a specific gap between efficient data-driven forecasting and physically reliable surrogate modeling. The proposed framework combines the speed and scalability of fully convolutional spatiotemporal learning with equation-based physical guidance derived from the CH dynamics. This design aims to preserve the practical efficiency needed for repeated forecasting while reducing the physical drift that can arise in purely data-driven long-horizon prediction. The objectives of this work are \textit{threefold}. First, we investigate whether CH-guided training improves the predictive fidelity of a fully convolutional spatiotemporal model for spinodal decomposition compared with a purely data-driven fully convolutional baseline. Second, we assess whether physics guidance improves long-horizon stability, especially when the model is used beyond its nominal prediction window through iterative rollout. Third, we evaluate whether the proposed model remains robust under reduced temporal context, where future evolution must be predicted from limited input observations. 
Numerical experiments demonstrate that the proposed physics-guided model consistently improves prediction performance over the purely data-driven baseline. The improvements are especially evident in extended-horizon and reduced-context settings, where the physics-guided model better preserves evolving morphology and reduces long-time drift. In addition to standard image-level metrics such as RMSE and SSIM, we evaluate physics-aware quantities including particle radius, particle count, interface curvature distribution, and tracked particle growth and shrinkage behavior. These results show that the benefit of physics guidance extends beyond visual similarity and leads to more faithful reproduction of physically meaningful coarsening trends.

The remainder of this paper is organized as follows. Section~2 introduces the CH phase-field formulation and residual formulation used for physics guidance. Section~3 presents the proposed physics-guided fully convolutional spatiotemporal learning framework. Section~4 describes the phase-field dataset, model training procedure, and evaluation metrics. Section~5 presents numerical results, including short-horizon prediction, long-horizon rollout, reduced-context forecasting, physics-based statistical evaluation, and computational efficiency. Section~6 discusses the implications and limitations of the proposed approach and concludes the paper with future research directions.

\section{Phase-Field Model and Residual Formulation}

In this work, we employ a CH phase-field formulation to model spinodal decomposition. The governing equation defines the physical evolution law of the concentration field and provides the basis for the physics-guided learning formulation. Rather than modifying the phase-field model, we reformulate the governing equation as a residual constraint that can be evaluated on predicted microstructure sequences during training. This section introduces the CH model, defines the spatiotemporal prediction notation, and presents the corresponding residual formulation of the governing equation.

\subsection{Cahn--Hilliard Model for Spinodal Decomposition}

Let $c(\mathbf{x},t)$ denote the local concentration of one component in a binary mixture, where $\mathbf{x}\in\Omega\subset\mathbb{R}^{2}$ and $t$ denotes time. In the CH formulation, $c$ is a conserved order parameter whose evolution is driven by reduction of the total free energy. The free-energy functional is written as
\begin{equation}
\mathcal{F}[c]
=
\int_{\Omega}
\left[
f_{\mathrm{chem}}(c)
+
\frac{\epsilon}{2}|\nabla c|^{2}
\right] d\mathbf{x},
\label{eq:free_energy}
\end{equation}
where $f_{\mathrm{chem}}(c)$ is the homogeneous chemical free-energy density, described by the regular solution model,  
\begin{equation}
f_{\mathrm{chem}}(c)
=
R\tilde{T}\left[c\ln c+(1-c)\ln(1-c)\right]
+
\omega c(1-c),
\label{eq:regular_solution}
\end{equation}
where $R$ is the gas constant, $\tilde{T}$ is the absolute temperature, and $\omega$ is the interaction parameter, and $\epsilon$ is the gradient-energy coefficient associated with diffuse interfaces. 
The corresponding chemical potential for Eq.(\ref{eq:free_energy}) is
\begin{equation}
\mu
=
\frac{\delta \mathcal{F}}{\delta c}
=
R\tilde{T}\ln\left(\frac{c}{1-c}\right)
+
\omega(1-2c)
-
\epsilon\nabla^{2}c .
\label{eq:chemical_potential}
\end{equation}
The concentration field evolves according to the CH equation
\begin{equation}
\frac{\partial c}{\partial t}
=
\nabla\cdot
\left[
M c(1-c)\nabla \mu
\right],
\label{eq:ch_equation}
\end{equation}
where $M$ is the mobility. Equivalently,
\begin{equation}
\frac{\partial c}{\partial t}
=
\nabla\cdot
\left\{
M c(1-c)
\nabla
\left[
R\tilde{T}\ln\left(\frac{c}{1-c}\right)
+
\omega(1-2c)
-
\epsilon\nabla^{2}c
\right]
\right\}.
\label{eq:expanded_ch}
\end{equation}
This equation describes chemical-potential-driven diffusion with concentration-dependent mobility. Under no-flux boundary conditions, the total concentration is conserved and the free energy is non-increasing in time. These properties are central to physically meaningful spinodal decomposition dynamics.

\subsection{Spatiotemporal Prediction Formulation and Physics-Guided Residual}
The phase-field simulation data are represented as time series concentration fields. Let
$X
=
\{x^{1},x^{2},\ldots,x^{T}\}$ 
denote an input sequence of observed microstructure fields, where each frame
$x^{t}\in\mathbb{R}^{H\times W}$ represents a discretized concentration field. Our objective is to predict the subsequent future sequence 
$Y 
=
\{x^{T+1},x^{T+2},\ldots,x^{T+T'}\}.$ 
The neural network predictor is written as
$\widehat{Y}
=
F_{\Theta}(X),
$
where $\Theta$ denotes the trainable parameters and
$\widehat{Y}
=
\{\widehat{x}^{T+1},\widehat{x}^{T+2},\ldots,\widehat{x}^{T+T'}\}$
is the predicted future sequence.

For the residual formulation, we use the simplified notation
\begin{equation}
\widehat{c}^{0}=x^{T},
\qquad
\widehat{c}^{k}=\widehat{x}^{T+k},
\quad k=1,\ldots,T'.
\end{equation}
To measure physical consistency, the CH residual is evaluated on the predicted sequence. The time derivative is approximated by a finite difference between consecutive frames,
\begin{equation}
D_{t}\widehat{c}^{k}
=
\frac{\widehat{c}^{k}-\widehat{c}^{k-1}}{\Delta t},
\label{eq:time_derivative}
\end{equation}
where $\Delta t$ is the time interval between stored simulation frames. 
The predicted chemical potential is computed as
\begin{equation}
\widehat{\mu}^{k}
=
R\tilde{T}\ln\left(\frac{\widetilde{c}^{k}}{1-\widetilde{c}^{k}}\right)
+
\omega(1-2\widetilde{c}^{k})
-
\epsilon \Delta_{h}\widehat{c}^{k},
\label{eq:predicted_mu}
\end{equation}
where $\Delta_{h}$ is the discrete Laplacian operator and $\widetilde{c}^{k}$ is a clipped version of $\widehat{c}^{k}$ used to avoid singular values in the logarithmic term. 
The discrete CH operator is given by
\begin{equation}
\mathcal{G}_{h}(\widehat{c}^{k})
=
\nabla_{h}\cdot
\left[
M\widetilde{c}^{k}(1-\widetilde{c}^{k})
\nabla_{h}\widehat{\mu}^{k}
\right],
\label{eq:ch_operator}
\end{equation}
where $\nabla_{h}$ and $\nabla_{h}\cdot$ denote the discrete gradient and divergence operators. The physics residual at the $k$-th predicted frame is then defined as
\begin{equation}
\mathcal{R}_{\mathrm{CH}}^{k}
=
D_{t}\widehat{c}^{k}
-
\mathcal{G}_{h}(\widehat{c}^{k}).
\label{eq:ch_residual}
\end{equation}
A smaller norm of $\mathcal{R}_{\mathrm{CH}}^{k}$ indicates that the predicted temporal evolution is more consistent with the governing CH dynamics. This residual is used to construct the physics-guided training objective for the fully convolutional spatiotemporal predictor in next section.

\section{Physics-Guided Fully Convolutional Spatiotemporal Learning Framework}

The residual formulation in Section~2 provides a way to evaluate whether a predicted microstructure sequence is consistent with the governing CH dynamics. In this section, we describe how this residual is used to train a fully convolutional spatiotemporal prediction model. The proposed framework retains the efficient direct-forecasting structure of fully convolutional spatiotemporal learning while using the CH residual as an additional physics-guided regularization term during training.

\subsection{Fully Convolutional Spatiotemporal Backbone}
The proposed model is built upon a fully convolutional encoder--translator--decoder architecture \cite{tan2022simvpv2}. Given an input sequence $X=\{x^{1},x^{2},\ldots,x^{T}\}$, the model predicts the future sequence 
\begin{equation}
\widehat{Y}=F_{\Theta}(X),
\label{eq:predseq}
\end{equation}
in a single forward pass. The network can be written abstractly as
\begin{equation}
\widehat{Y}
=
D_{\Theta}
\left(
P_{\Theta}
\left(
E_{\Theta}(X)
\right)
\right),
\label{eq:encoder_translator_decoder}
\end{equation}
where $E_{\Theta}$ denotes the spatial encoder, $P_{\Theta}$ denotes the temporal translator, and $D_{\Theta}$ denotes the spatial decoder. The spatial encoder maps the input microstructure sequence into a latent feature representation. Its role is to extract local morphological features from the concentration fields, including phase interfaces, concentration gradients, domain boundaries, and spatial connectivity patterns. Since the encoder is fully convolutional, it operates directly on dense phase-field images and is not tied to a fixed spatial resolution. The temporal translator models the evolution of the latent microstructure features. Instead of propagating a recurrent hidden state frame by frame, the translator uses stacked convolutional spatiotemporal blocks to transform the encoded representation into future latent states. This non-recurrent design enables efficient multi-step prediction and avoids the sequential inference cost associated with recurrent forecasting models. Within the temporal translator, gated spatiotemporal attention-type (gSTA) blocks are used to capture evolving morphology. Depth-wise spatial convolutions provide efficient local feature extraction, dilated convolutions enlarge the receptive field to represent longer-range interactions associated with diffusion and coarsening, and $1\times1$ channel-wise convolutions mix information across latent feature channels. The gating mechanism adaptively regulates the contribution of evolving latent features, allowing the model to emphasize regions undergoing active morphological change while preserving relatively stable regions.

\begin{figure}[t]
\centering
\includegraphics[width=\linewidth, trim=0cm 0cm 9cm 0cm]{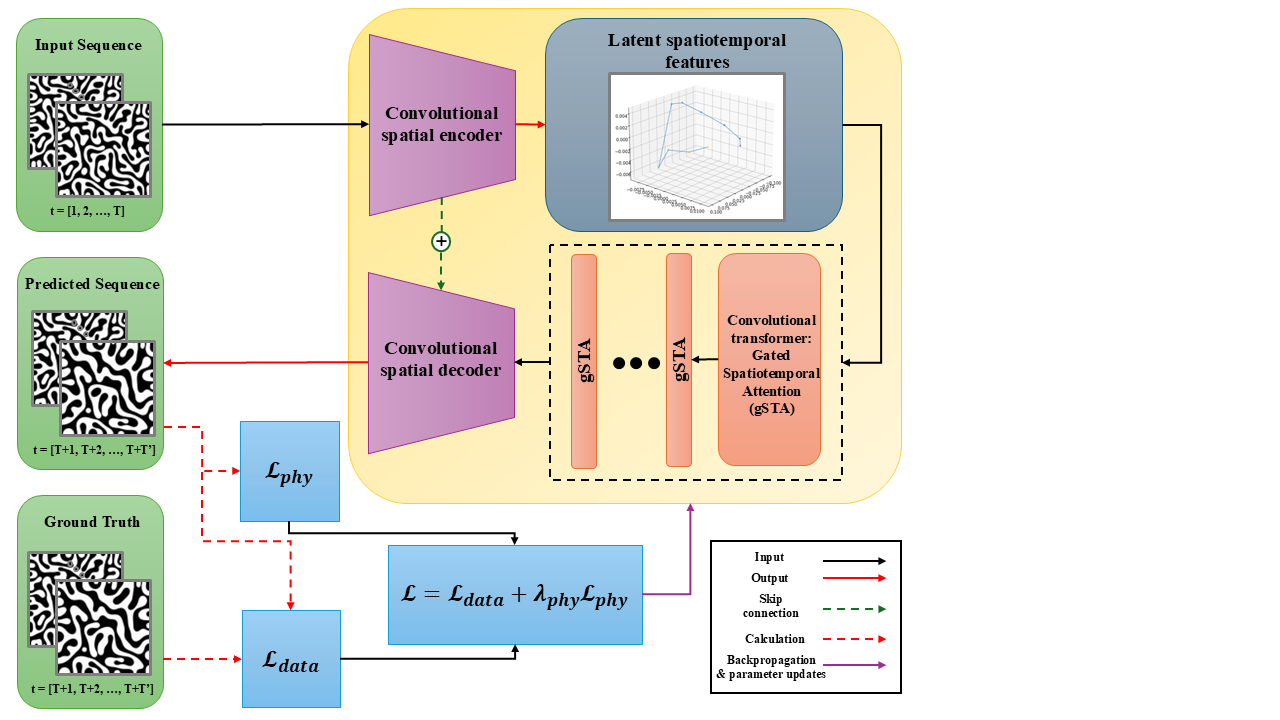}
\caption{Flowchart of the proposed physics-guided fully convolutional spatiotemporal framework for microstructure evolution prediction. An input sequence of length $T$ is first processed by a convolutional spatial encoder to extract microstructural spatial correlations and project the data into a compact latent feature space. The latent representations are then propagated through stacked gated Spatiotemporal Attention (gSTA) modules, which serve as the temporal translator of the network and model the dynamic evolution patterns of the microstructure. A physics-guided regularization term is incorporated during training to constrain the predicted evolution to remain consistent with the governing phase-field dynamics. Finally, a convolutional spatial decoder reconstructs the forecast microstructure fields at future time steps $T+1$ through $T+T'$.}
\label{flowchartNew}
\end{figure}

Finally, the spatial decoder maps the translated latent representation back to the physical image space and reconstructs the predicted future concentration fields. Because all components are convolutional, the model can be evaluated on spatial resolutions different from those used during training, without changing the network architecture. The fully convolutional spatiotemporal network architecture is presented in the flowchart, see Figure \ref{flowchartNew}.

\subsection{Physics-Guided Training Objective}

The central difference between the proposed model and a purely data-driven fully convolutional predictor lies in the training objective. A purely data-driven model is trained only to minimize the discrepancy between predicted and reference future microstructure fields. In contrast, the proposed model is trained using both data supervision and CH-based physical regularization. Using the notation introduced in Section~2, let $\widehat{c}^{k}=\widehat{x}^{T+k}$ denote the predicted concentration field and let
\begin{equation}
c^{k}=x^{T+k},
\qquad
k=1,\ldots,T',
\end{equation}
denote the corresponding reference concentration field. The data-fitting loss is defined as
\begin{equation}
\mathcal{L}_{\mathrm{data}}
=
\frac{1}{T'|\Omega_{h}|}
\sum_{k=1}^{T'}
\left\|
\widehat{c}^{k}
-
c^{k}
\right\|_{2}^{2},
\label{eq:data_loss}
\end{equation}
where $|\Omega_{h}|=H\times W$ is the number of spatial grid points. The physics-guided loss is constructed from the CH residual defined in Eq.~\eqref{eq:ch_residual}. Specifically,
\begin{equation}
\mathcal{L}_{\mathrm{phy}}
=
\frac{1}{T'|\Omega_{h}|}
\sum_{k=1}^{T'}
\left\|
\mathcal{R}_{\mathrm{CH}}^{k}
\right\|_{2}^{2}.
\label{eq:physics_loss}
\end{equation}
This term penalizes predicted sequences whose temporal evolution deviates from the governing CH dynamics.
The total loss used to train the physics-guided model is
\begin{equation}
\mathcal{L}
=
\mathcal{L}_{\mathrm{data}}
+
\lambda_{\mathrm{phy}}
\mathcal{L}_{\mathrm{phy}},
\label{eq:total_loss}
\end{equation}
where $\lambda_{\mathrm{phy}}$ controls the relative weight of the physics-guided regularization term. 
The data-fitting term encourages agreement with the reference phase-field simulations, while the physics-guided term encourages the predicted sequence to follow mass-conserving, chemical-potential-driven phase separation dynamics. Thus, the model is trained not only to reproduce future microstructure fields, but also to generate trajectories that are more consistent with the underlying physical evolution law.

\subsection{Training and Inference Strategy}

During training, the input sequence is first passed through the fully convolutional predictor to obtain the complete predicted future sequence $\widehat{Y}$. 
The data-fitting loss is computed by comparing the predicted sequence with the reference future sequence. In parallel, the CH residual defined in Eq.~\eqref{eq:ch_residual} is evaluated on the predicted concentration fields using the finite-difference temporal derivative and discrete spatial operators described in Section~2. The total loss in Eq.~\eqref{eq:total_loss} is then minimized with respect to the network parameters $\Theta$. The same fully convolutional architecture is used for both the purely data-driven baseline and the proposed physics-guided model. The baseline model is trained using only $\mathcal{L}_{\mathrm{data}}$, while the physics-guided model is trained using $\mathcal{L}$. This design allows the effect of CH-based physical regularization to be isolated and evaluated directly.

Importantly, the CH residual is evaluated only during training. The governing equation is not solved inside the training loop, and no additional phase-field simulation is performed. Therefore, the physics-guided term acts as an equation-based regularizer for the learned spatiotemporal mapping rather than as a replacement for the neural-network predictor. At inference time, the model predicts future microstructure fields through the same direct fully convolutional forward pass as Eq.~\eqref{eq:predseq}. 
No CH residual is evaluated during inference, no CH equation is solved, and no additional optimization problem is introduced. As a result, the proposed method preserves the computational efficiency of fully convolutional spatiotemporal forecasting.

Overall, the proposed framework combines an efficient fully convolutional spatiotemporal predictor with CH-based physical regularization during training. This design aims to improve prediction fidelity, reduce long-horizon drift, and better preserve physically meaningful coarsening behavior while maintaining fast inference.

\section{Experimental Setting}
\label{sec:exp}

\subsection{Phase-Field Simulation Dataset}

The training and evaluation data consist of previously published phase-field simulation datasets \cite{yang2021self}, which provide ground-truth microstructure evolution for spinodal decomposition. To verify robustness and reproducibility, additional simulations are performed under identical physical and numerical settings. No modifications to the governing equation, model parameters, or numerical solver are introduced in this work. Phase-field methods provide a diffuse-interface framework for modeling microstructure evolution. In these models, phase concentrations are represented by continuous order parameters whose dynamics are governed by thermodynamically derived PDEs. This formulation naturally captures interfacial motion, topological transitions, and long-time morphological evolution without explicit interface tracking.

Spinodal decomposition is simulated using the CH formulation described in Section~2. In particular, the simulations considered in this work focus on the second coarsening stage of phase separation, which is the regime used in the published benchmark dataset \cite{yang2021self}. The dimensionless parameters are set to \(R\tilde{T}=1\), \(\omega = 0.27397\), \(\epsilon = 0.1682\), and \(M = 1\), with spatial grid spacing \(\Delta x = \Delta y = 1\) in all simulations. The governing equation is solved using an implicit variable-order backward differentiation formula (BDF) solver implemented in COMSOL Multiphysics \cite{multiphysics1998introduction}. The average dimensionless time-step size is 4.01. Simulation outputs are recorded every 1,500 dimensionless time units, corresponding to approximately 370 adaptive BDF solver steps between consecutive stored frames. The resulting concentration fields are used as image sequences representing the temporal evolution of phase separation.

\subsection{Dataset Construction and Preprocessing}
\label{sec:DataPre}

All spatiotemporal microstructure sequences are organized as tensors of shape 
$
\mathbb{R}^{B \times T \times C \times H \times W},
$
where $B$ denotes the number of samples, $T$ is the temporal length, $C$ is the number of channels, and $H$ and $W$ are the spatial dimensions. In this work, all microstructure fields are represented as single-channel grayscale images, so $C=1$. 
For each training sample, we use $10$ consecutive frames as input and predict the subsequent $T'=90$ frames, resulting in a 100-frame spatiotemporal window. Training clips are generated from each long simulation trajectory using a sliding-window procedure. If a sequence contains $T_{\mathrm{seq}}$ frames and the stride is $S$, then the number of generated clips is
$
\left\lfloor \frac{T_{\mathrm{seq}}-100}{S} \right\rfloor + 1.
$
This procedure enables efficient reuse of long trajectories while preserving temporal continuity. In all sliding-window experiments, the stride was set to $S=10$.

To evaluate robustness under reduced temporal information, we also consider prediction from a single static input frame. Since the model is trained with a 10-frame input interface, zero-padding is applied to the beginning of the input sequence when fewer than 10 frames are available. For example, in the 1$\rightarrow$99 setting, one observed frame is padded with nine leading zero frames before being passed to the network. This allows the same trained architecture to be used for reduced-context inference without retraining. 
To extend prediction beyond the nominal 90-frame output horizon, iterative roll-out is employed. Specifically, the final 10 predicted frames are recursively fed back into the model as input to generate subsequent predictions. This procedure is used for the 10$\rightarrow$200 forecasting experiments.

All datasets are partitioned into disjoint training, validation, and test sets at the trajectory level to avoid temporal information leakage. For spinodal decomposition, the training set contains approximately 580 trajectories and the validation set contains 40 trajectories, both at $64\times64$ resolution. Applying the sliding-window procedure yields 6,380 training clips and 440 validation clips. The test set contains 510 independent trajectories at $64\times64$ resolution and 50 independent trajectories at $256\times256$ resolution, both containing 210 frames. These higher-resolution sequences are used to assess both predictive accuracy and resolution-transfer capability.

\subsection{Physics-Guided Fully Convolutional Model Training and Inference}

The proposed model is built upon a fully convolutional spatiotemporal architecture with physics-guided regularization. The spatial encoder and decoder each contain four spatial attention blocks with 64 hidden channels, while the temporal translator contains eight temporal attention blocks with 256 hidden channels. Greater capacity is assigned to the temporal translator because long-horizon microstructure prediction requires the model to capture accumulated nonlinear spatiotemporal dynamics. The model is trained for up to 200 epochs with batch size 1 using the Adaptive Moment Estimation (Adam) optimizer with an initial learning rate of \(10^{-3}\). The physics-guided regularization weight is fixed as \(\lambda_{\mathrm{phy}}=0.01\) in all reported experiments. While this auxiliary weight can, in principle, be treated as a learnable parameter, we fix it here for stability and empirical effectiveness.

Training is performed on the \(64\times 64\) spinodal decomposition dataset using the AI Panther system at the Florida Institute of Technology, equipped with NVIDIA A100 SXM4 GPUs. In addition to the baseline fully convolutional model trained without physics guidance, we train a physics-guided version in which the governing CH dynamics are incorporated into the learning objective through a residual regularization term. For the physics-guided model, the CH residual is evaluated directly on the predicted concentration fields during training. In this residual evaluation, we use a normalized frame interval with \(\Delta t=1\), so that one unit of time corresponds to the spacing between two adjacent stored simulation frames. This normalization keeps the residual evaluation consistent with the discrete temporal indexing of the training sequences.

The discrete operators \(\Delta_h\), \(\nabla_h\), and \(\nabla_h\cdot\) in Eqs.~\eqref{eq:predicted_mu} and \eqref{eq:ch_operator} are implemented using finite-difference convolution kernels, enabling efficient residual evaluation within the neural-network training loop. Replication padding is used in the finite-difference convolution kernels to approximate homogeneous Neumann boundary conditions, consistent with the no-flux phase-field simulations. To avoid numerical singularities in the logarithmic chemical-potential term, the predicted concentration field is clipped only when evaluating the logarithmic and mobility terms. Specifically, the clipped concentration field is defined as
$\widetilde{c}^{k}=\mathrm{clip}(\widehat{c}^{k},\delta,1-\delta)$, 
where \(\delta=10^{-4}\) in our experiments. The Laplacian term \(\Delta_h \widehat{c}^{k}\) is evaluated using the unclipped predicted concentration field \(\widehat{c}^{k}\). Thus, clipping is used only for numerical stability in the nonlinear chemical-potential and mobility terms and does not directly modify the interfacial-energy contribution.

No CH solver is used during either training or inference; the residual serves only as an equation-based regularization term for training the neural-network predictor. At inference time, the trained model predicts future microstructure fields through a direct fully convolutional forward pass. For computational comparison, we also report the performance of E3D-LSTM under the same benchmark setting. After training on \(64\times64\) data, the models are directly evaluated on \(256\times256\) sequences without retraining. Owing to the fully convolutional design, no architectural modification is required for higher-resolution inference. This setup enables direct assessment of spatial-resolution transfer, prediction accuracy, and computational efficiency.

\subsection{Performance Metrics}

Prediction accuracy is first assessed using standard image-based metrics, namely RMSE and SSIM \cite{ZWang04}. Both metrics are evaluated every 5 frames over the prediction horizon. Because pixel-level agreement alone is insufficient to assess the physical fidelity of spinodal decomposition, we additionally evaluate several physics-based statistical quantities. These include the average particle radius, particle count, interface-segment curvature distribution, and tracked particle growth/shrinkage behavior. These metrics are used to assess whether the model preserves physically meaningful coarsening trends and interfacial evolution beyond image similarity. For visualization consistency, all predicted frames are displayed using a fixed intensity range; this normalization is used only for visualization and does not affect metric computation.

\section{Numerical Results}
\label{sec:res}
 
In this section, we evaluate the proposed physics-guided fully convolutional spatiotemporal framework for spinodal decomposition. The experiments are designed to assess both short-term forecasting accuracy and long-horizon extrapolation capability, as well as the preservation of physics-based statistical properties. We compare the results against a baseline fully convolutional spatiotemporal framework (SimVPv2 without physics guidance) and an E3D-LSTM model trained and tested on the same datasets. Spinodal decomposition describes the spontaneous phase separation of a homogeneous mixture into compositionally distinct domains. In this work, the ground-truth evolution is generated by solving the CH equation~\eqref{eq:ch_equation}, which couples interfacial energy with long-range diffusion and yields fourth-order nonlinear dynamics. Hence, this diffusion-coupled evolution is more challenging to forecast, particularly over long horizons. The process typically exhibits a rapid early-time amplification of composition fluctuations followed by a slower coarsening stage, and we therefore assess prediction performance across both early and late evolution periods.

\subsection{Predictions of 90 future frames from 10 input frames (10$\rightarrow$90) with resolution $64 \times 64$}

\begin{figure}
\centering
\includegraphics[width=\linewidth]{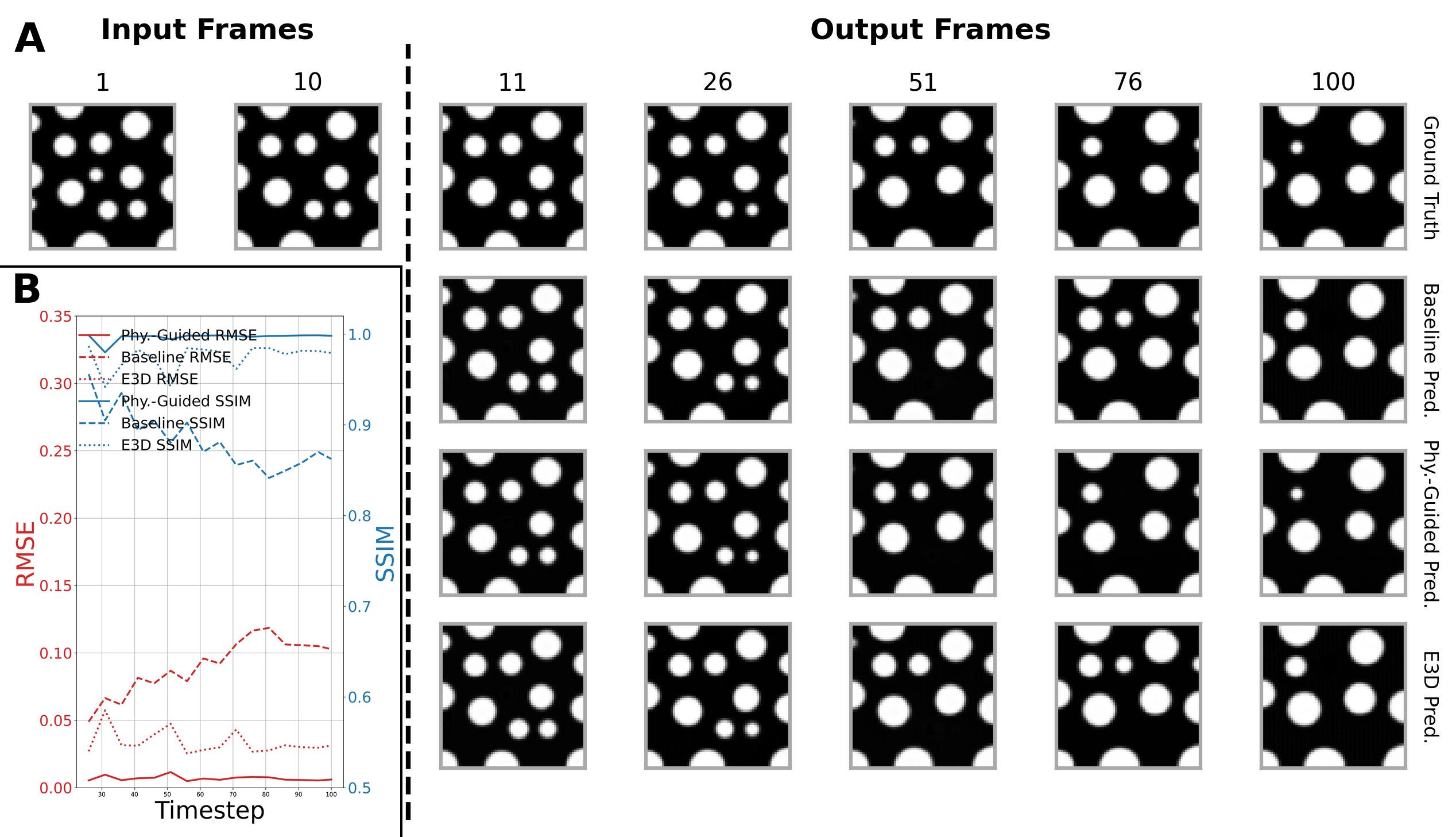}
\caption{\textbf{Spinodal decomposition prediction (10 input frames, 90 output frames) comparison:} Two sets of predictions are displayed along the ground truth. (A) Prediction sequence and the corresponding ground-truth sequence, shown at t = 11, 26, 51, 76, 100; (B) RMSE and SSIM computed between the predicted and ground-truth sequences at 5-frame intervals.}
\label{fig4}
\end{figure}

\begin{figure}[h!]
\centering
\includegraphics[width=\linewidth]{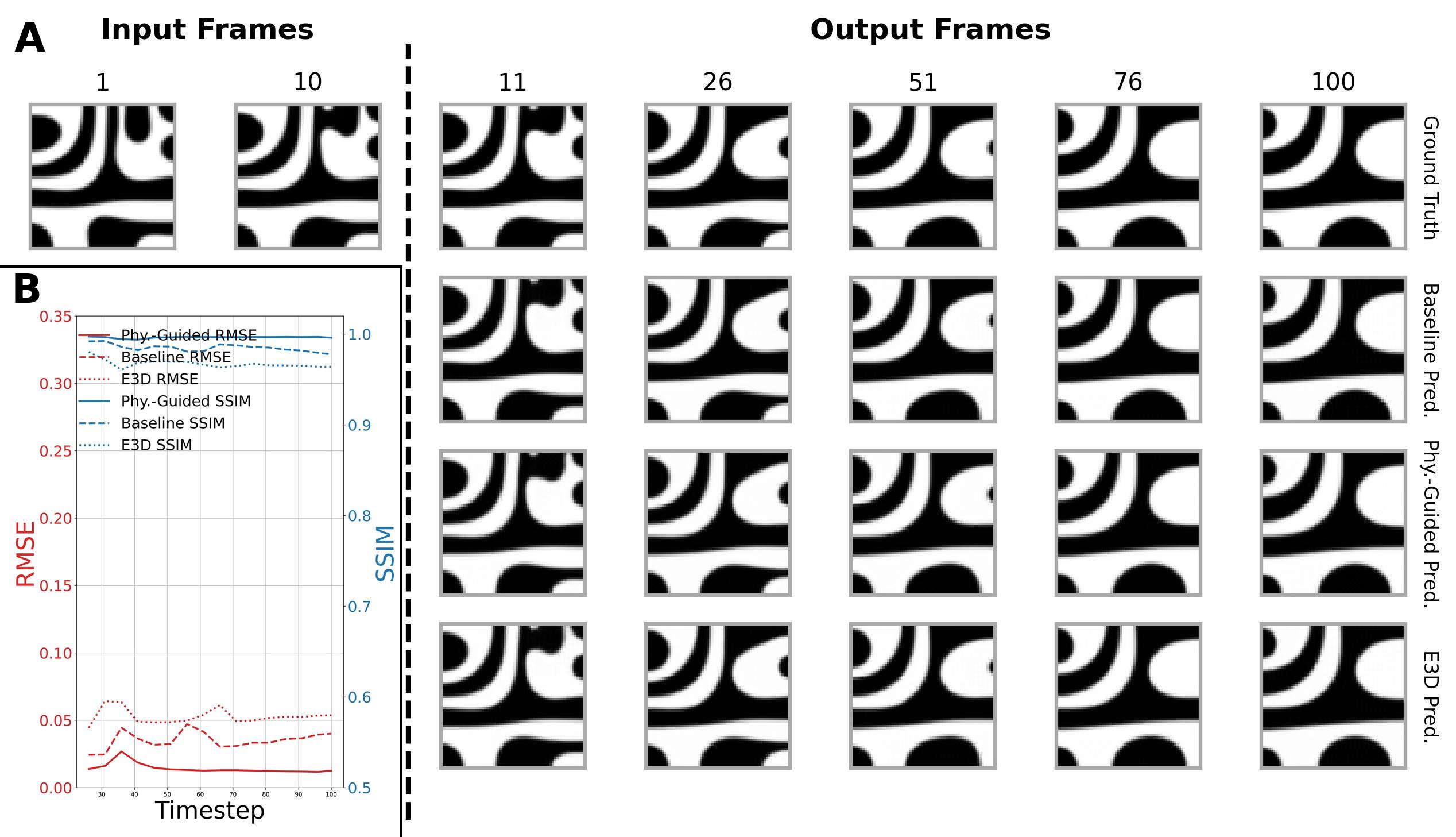}
\caption{\textbf{Spinodal decomposition prediction (10 input frames, 90 output frames) comparison:} Two sets of predictions are displayed along the ground truth. (A) Prediction sequence and the corresponding ground-truth sequence, shown at t = 11, 26, 51, 76, 100; (B) RMSE and SSIM computed between the predicted and ground-truth sequences at 5-frame intervals.}
\label{fig5}
\end{figure}

For the 10$\rightarrow$90 task, 1000 test samples are selected from the 6,120 available samples for evaluation. Two representative examples are shown in Figs.~\ref{fig4} ($c_{0}$ = 0.25) and~\ref{fig5} ($c_{0}$ = 0.5). The input sequence is displayed at $t=1$ and $t=10$, while the prediction and ground-truth sequences are shown at $t=11, 26, 51, 76,$ and $100$ (Figs.~\ref{fig4}.A and \ref{fig5}.A). The prediction sequences include the baseline fully convolutional spatiotemporal network, the physics-guided fully convolutional spatiotemporal network, and the E3D-LSTM network. Based on these visual results, the physics-guided network captures particle morphology with higher fidelity than the other two models. Bicontinuous phase separation dynamics are comparable across all three models. Quantitative metrics (RMSE and SSIM), computed at 5-frame intervals and shown in Figs.~\ref{fig4}.B and \ref{fig5}.B, further support this observation, with the physics-guided network achieving the best performance across both concentration cases. Despite its strong performance in forecasting bicontinuous phase separation dynamics, the baseline network exhibits limited capability in capturing particle morphology.

\begin{figure}[h!]
\centering
\includegraphics[width=0.6\linewidth]{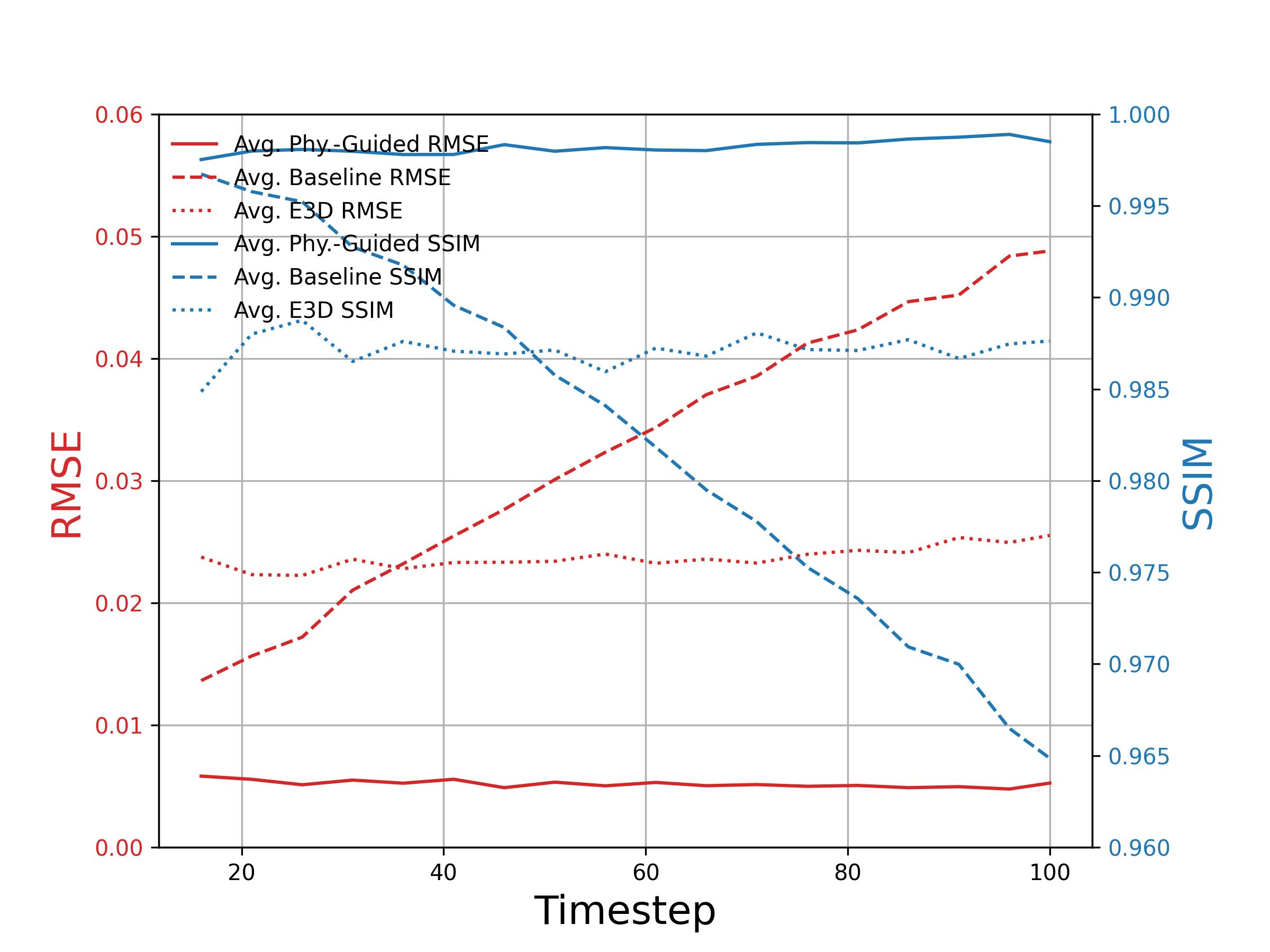}
\caption{\textbf{Accuracy metrics of physics-guided model and baseline model for spinodal decomposition prediction (10 input frames, 90 output frames):} Dataset-averaged RMSE and SSIM evaluated every 5 frames over the prediction horizon. Physics-guided model metrics are represented by the solid line, the baseline model metrics are represented by the dashed line, and the E3D-LSTM model metrics are represented by the dotted line.}
\label{fig6}
\end{figure}

Average RMSE and SSIM values across all 1000 testing samples are shown in Fig.~\ref{fig6} for the three trained networks. The physics-guided network achieves the best overall performance, with E3D-LSTM exhibiting comparable results. The baseline network performs well for short-term prediction but gradually loses accuracy over longer horizons. In contrast, the physics-guided network maintains strong predictive performance while retaining relatively low computational complexity, providing an effective balance between efficiency and spatiotemporal accuracy.



\begin{figure}[h!]
\centering
\includegraphics[width=\linewidth]{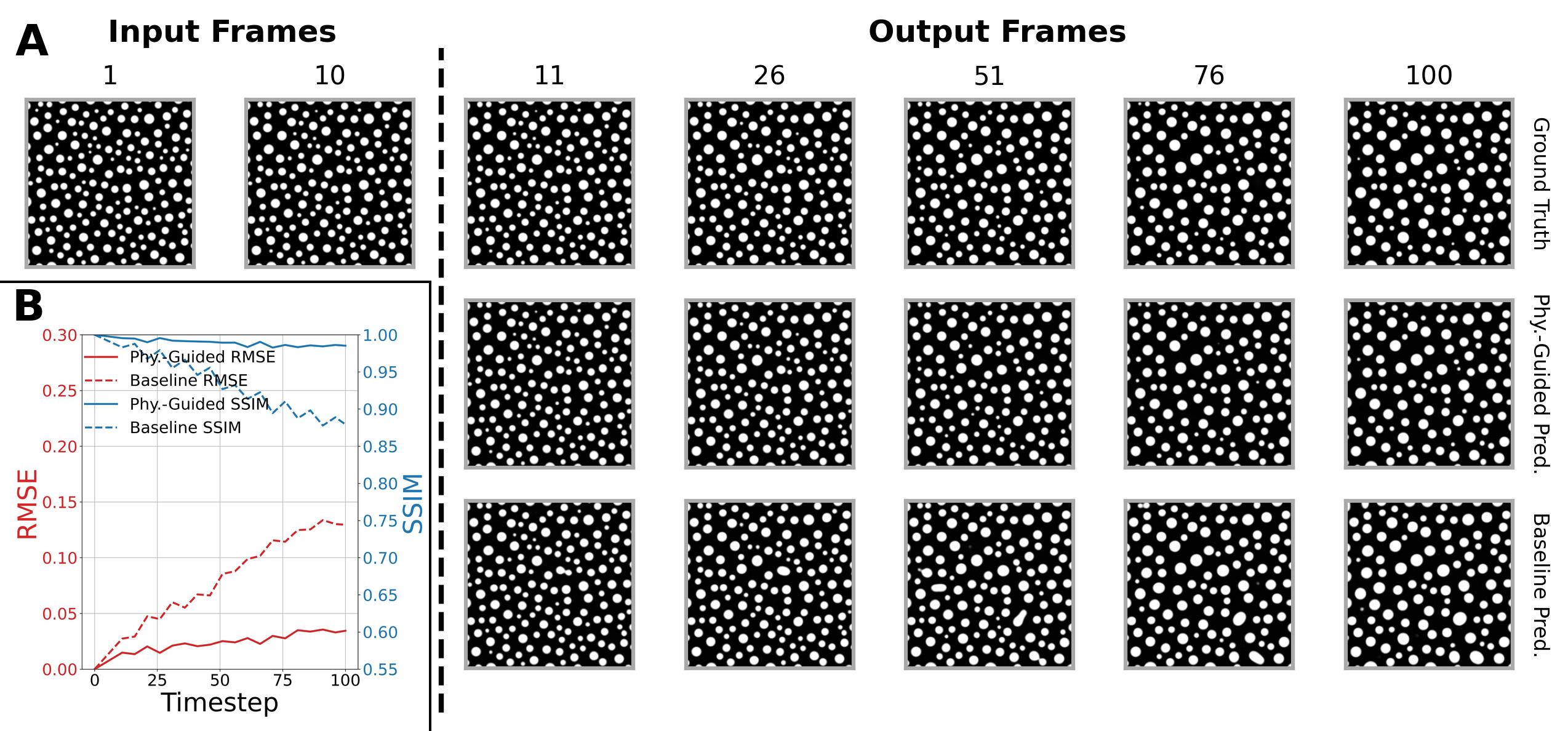}
\caption{\textbf{Spinodal decomposition prediction (10 input frames, 90 output frames) comparison:} Two sets of predictions are displayed along the ground truth. (A) Prediction sequence and the corresponding ground-truth sequence, shown at t = 11, 26, 51, 76, 100; (B) RMSE and SSIM computed between the predicted and ground-truth sequences at 5-frame intervals.}
\label{fig7}
\end{figure}

\begin{figure}[h!]
\centering
\includegraphics[width=\linewidth]{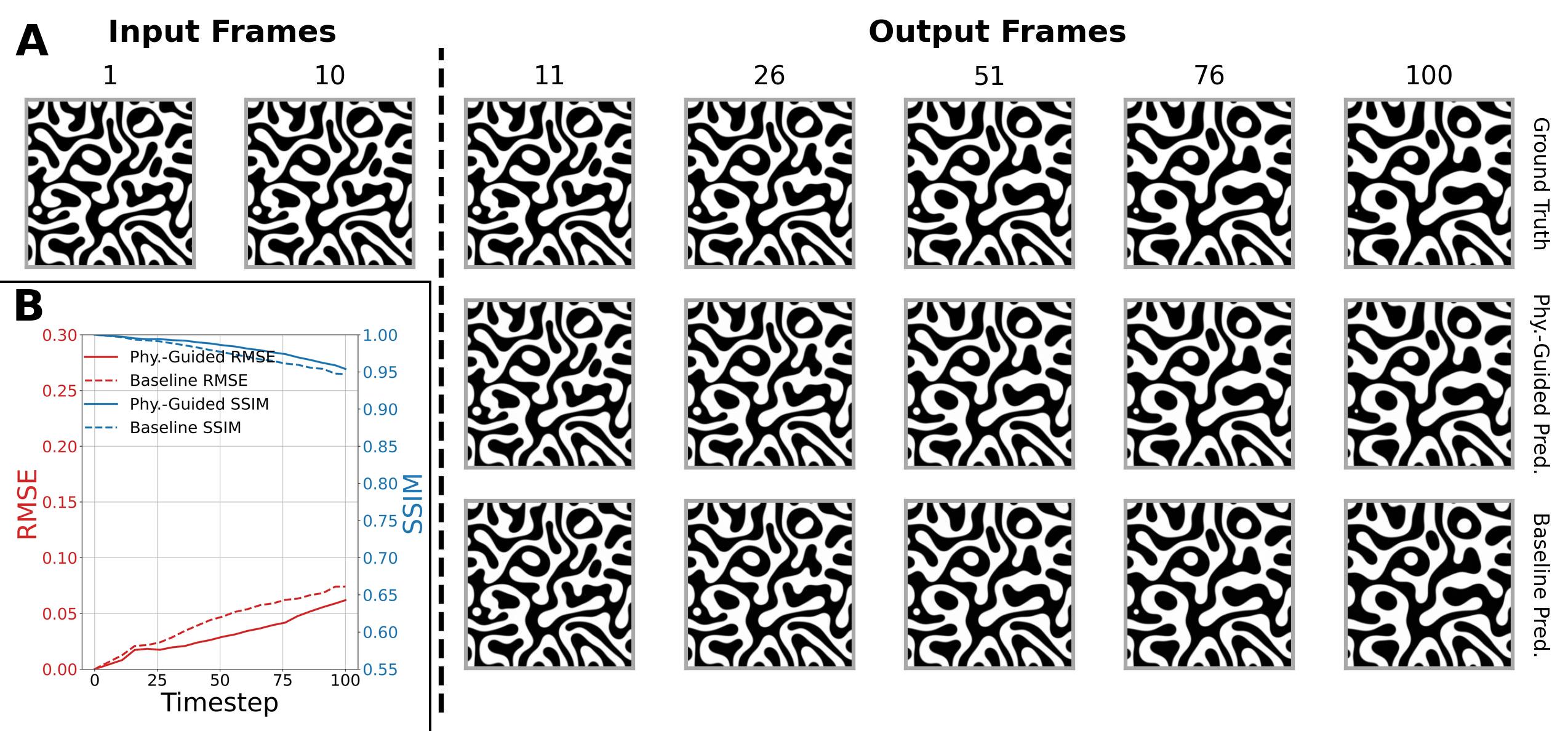}
\caption{\textbf{Spinodal decomposition prediction (10 input frames, 90 output frames) comparison:} Two sets of predictions are displayed along the ground truth. (A) Prediction sequence and the corresponding ground-truth sequence, shown at t = 11, 26, 51, 76, 100; (B) RMSE and SSIM computed between the predicted and ground-truth sequences at 5-frame intervals.}
\label{fig8}
\end{figure}

\subsection{Predictions of 90 future frames from 10 input frames (10$\rightarrow$90) with resolution $256 \times 256$}

For the 10$\rightarrow$90 task, 600 test samples are evaluated. Two representative examples are shown in Figs.~\ref{fig7} ($c_{0}$ = 0.25) and~\ref{fig8} ($c_{0}$ = 0.5). The input sequence is displayed at $t=1$ and $t=10$, while the predicted and ground-truth sequences are shown at $t=11, 26, 51, 76,$ and $100$ (Figs.~\ref{fig7}.A and \ref{fig8}.A). The prediction sequences include the baseline fully convolutional spatiotemporal network, and the physics-guided fully convolutional spatiotemporal network. In Fig.~\ref{fig7}.A, droplet domains are well localized in both models. However, the physics-guided network more accurately captures droplet morphology evolution than the baseline network. This improvement is further supported by the quantitative metrics shown in Fig.~\ref{fig7}.B. Additionally, the physics-guided network better preserves morphology near the domain boundary. Fig.~\ref{fig8}.B shows a similar trend, with the physics-guided model producing slightly finer phase-separation forecasts. Although the baseline model performs comparably in predicting bicontinuous phase distributions, the physics-guided model remains more robust across varying concentration regimes.

\begin{figure}[h!]
\centering
\includegraphics[width=0.6\linewidth]{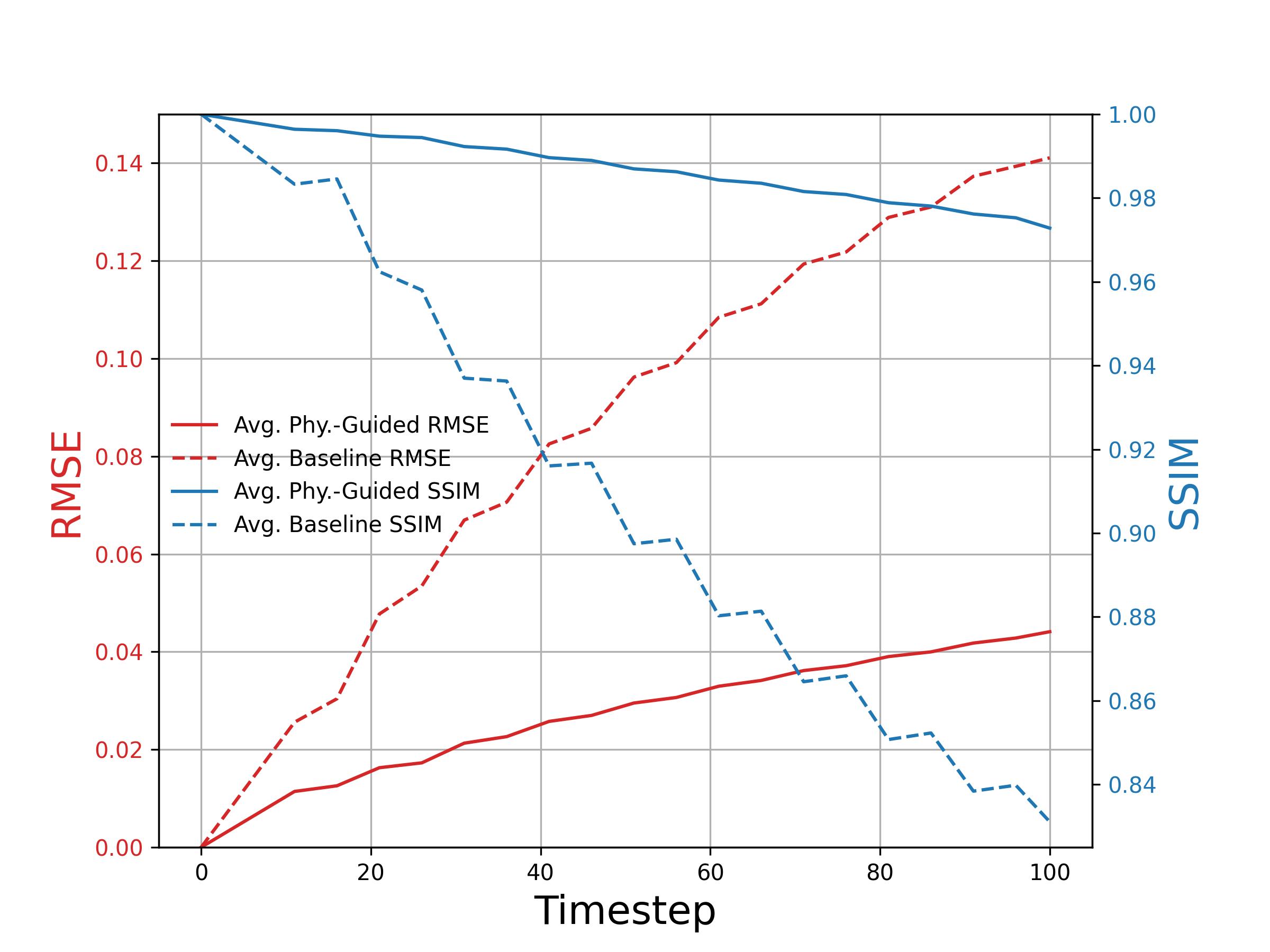}
\caption{\textbf{Accuracy metrics of physics-guided model and baseline model for spinodal decomposition prediction (10 input frames, 90 output frames):} Dataset-averaged RMSE and SSIM evaluated every 5 frames over the prediction horizon. Physics-guided model metrics are represented by the solid line, and the baseline model metrics are represented by the dashed line.}
\label{fig9}
\end{figure}

\begin{figure}[h!]
\centering
\includegraphics[width=\linewidth]{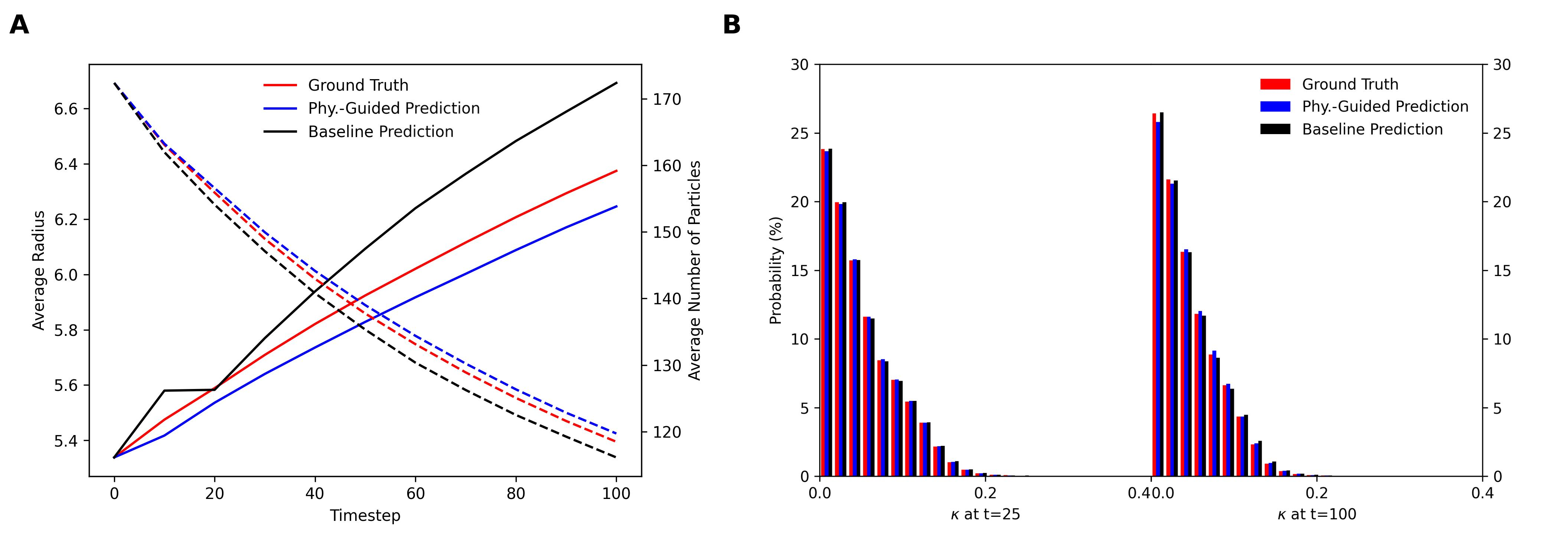}
\caption{\textbf{Physics-based metrics of physics-guided model and baseline model for spinodal decomposition prediction (10 input frames, 90 output frames):} (A) Average radius and average number of particles across $c_{0}=0.25$ samples; (B) Interface segment curvature distribution at timesteps t=25, 100 for samples with concentration $c_{0}=0.5$.}
\label{fig10}
\end{figure}

Dataset-level performance is summarized in Fig.~\ref{fig9}. The mean RMSE and SSIM, computed at 5-frame intervals, indicate consistently strong predictive performance across the dataset. Physics-based evaluation results are presented in Fig.~\ref{fig10}. In Fig.~\ref{fig10}.A, the average particle radius and particle count are shown for all test samples with concentration $c_{0}=0.25$. The physics-guided network preserves these properties with higher fidelity than the baseline network, demonstrating the benefit of physics-guided regularization. Fig.~\ref{fig10}.B shows the interface curvature distribution for samples with $c_{0}=0.5$ at timesteps $t = 25$ and $t = 100$. Both models closely match the ground-truth distribution, consistent with the strong performance of the baseline network in predicting phase separation dynamics and phase distributions.

\begin{figure}[h!]
\centering
\includegraphics[width=0.7\linewidth]{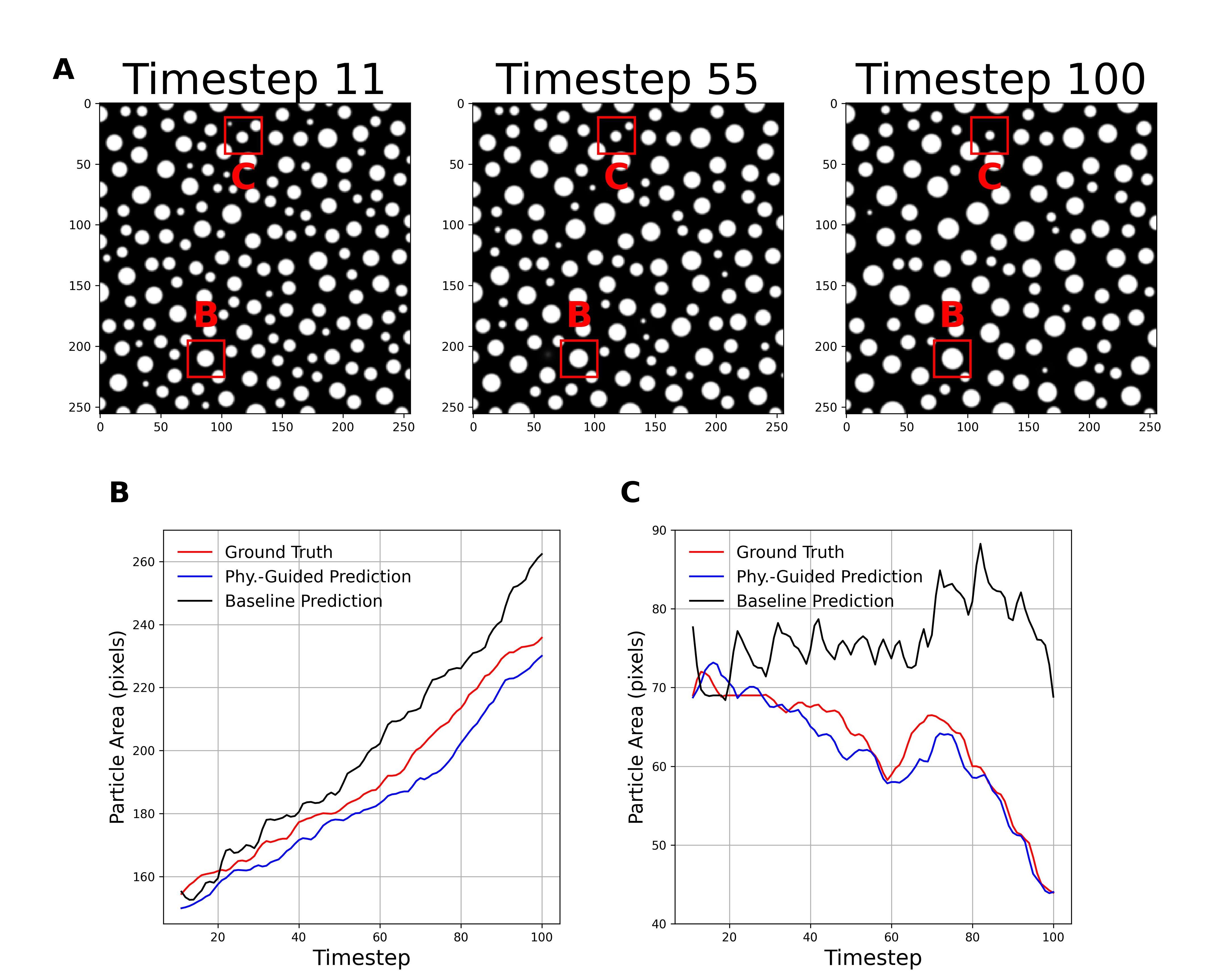}
\caption{\textbf{Tracked particle growth/shrink of physics-guided model and baseline model for spinodal decomposition prediction (10 input frames, 90 output frames):} (A) Tracked particle subjects; (B) growth of tracked particle on ground truth, physics-guided model, and baseline model; (C) shrinking of tracked particle on ground truth, physics-guided model, and baseline model.}
\label{fig11}
\end{figure}

Fig.~\ref{fig11} compares the temporal evolution of growing and shrinking particles between model predictions and the ground truth sequence. The ground truth frames, shown at $t = 11, 55,$ and $100$ in Fig.~\ref{fig11}.A, correspond to the same sequence presented in Fig.~\ref{fig7}. Selected particles are highlighted, with the growing and shrinking cases tracked in Fig.~\ref{fig11}.B and Fig.~\ref{fig11}.C, respectively. Particle area is quantified at each timestep and smoothed using a Savitzky–Golay filter for improved interpretability. The physics-guided network more closely follows the ground truth during both growth and shrinkage, while exhibiting reduced fluctuations in particle size compared to the baseline network. These results highlight the benefit of incorporating physics-based regularization into the model.

\begin{figure}[h!]
\centering
\includegraphics[width=\linewidth]{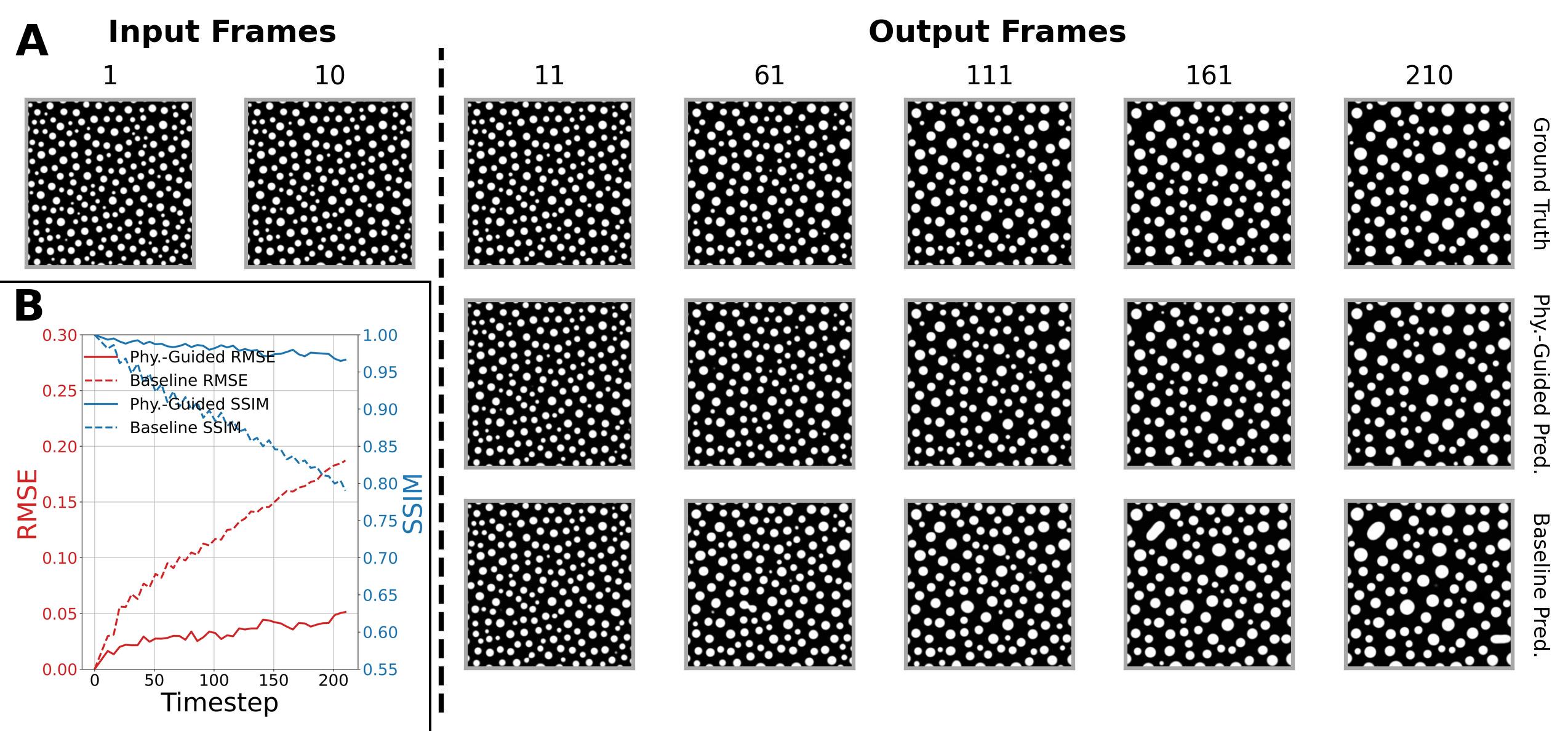}
\caption{\textbf{Spinodal decomposition prediction (10 input frames, 200 output frames) comparison:} Two sets of predictions are displayed along the ground truth. (A) Prediction sequence and the corresponding ground-truth sequence, shown at t = 11, 61, 111, 161, 210; (B) RMSE and SSIM computed between the predicted and ground-truth sequences at 5-frame intervals.}
\label{fig12}
\end{figure}

\begin{figure}[h!]
\centering
\includegraphics[width=\linewidth]{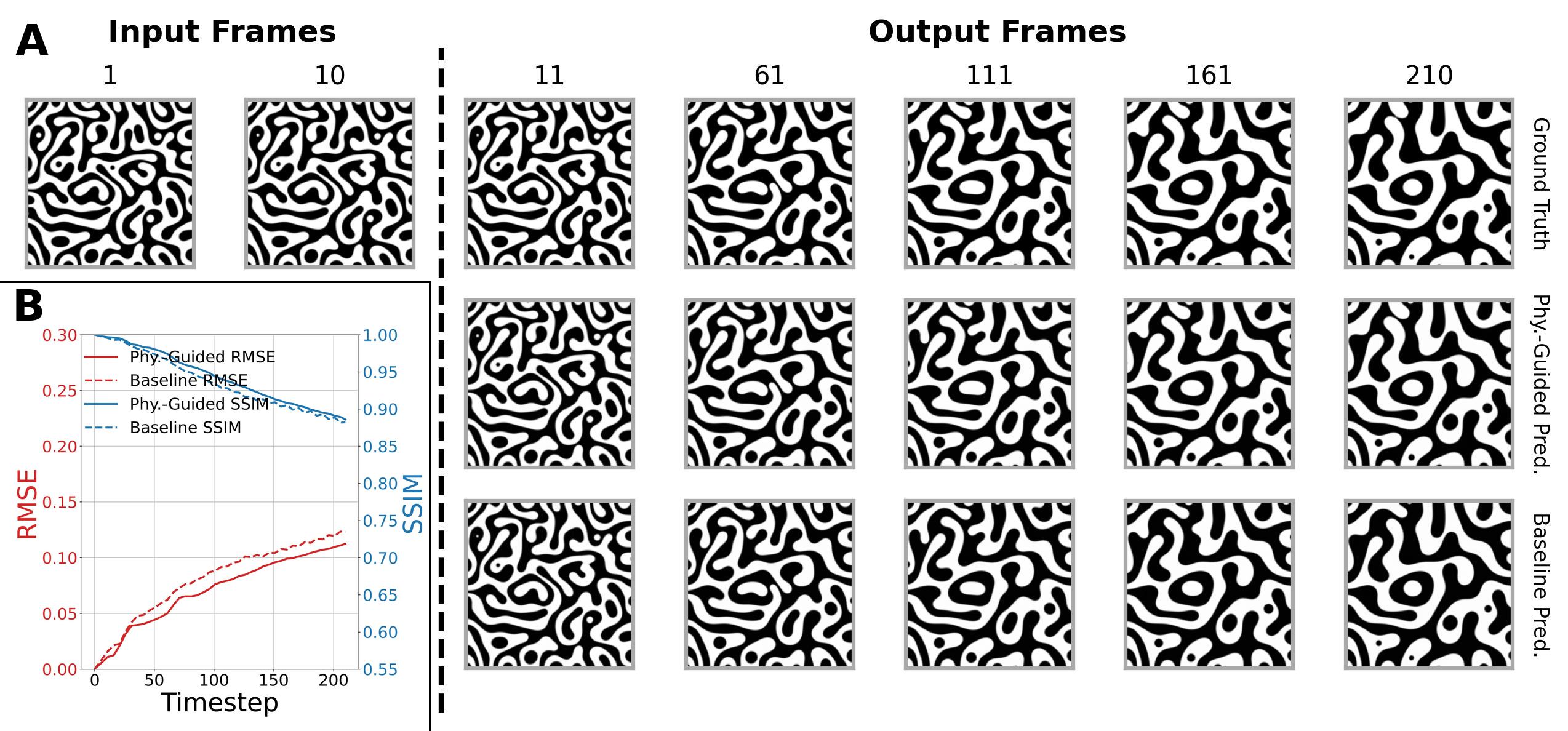}
\caption{\textbf{Spinodal decomposition prediction (10 input frames, 200 output frames) comparison:} Two sets of predictions are displayed along the ground truth. (A) Prediction sequence and the corresponding ground-truth sequence, shown at t = 11, 61, 111, 161, 210; (B) RMSE and SSIM computed between the predicted and ground-truth sequences at 5-frame intervals.}
\label{fig13}
\end{figure}

\subsection{Predictions of 200 future frames from 10 input frames (10$\rightarrow$200)}

For the 10$\rightarrow$200 task, 50 test samples are evaluated on the network models that are trained to predict 90 future frames from 10 input frames. Using iterative roll-out, we can extend the prediction horizon beyond the nominal 90-frame output length. Two representative examples are shown in Figs.~\ref{fig12} ($c_{0}$ = 0.25) and~\ref{fig13} ($c_{0}$ = 0.5). For compact visualization, the input sequence is displayed at $t=1$ and $t=10$, while the predicted and ground-truth outputs are shown at $t=11, 61, 111, 161,$ and $210$ (Figs.~\ref{fig12}.A and \ref{fig13}.A). The prediction sequences include the baseline fully convolutional spatiotemporal network, and the physics-guided fully convolutional spatiotemporal network. Fig.~\ref{fig12}.A shows improved droplet spatiotemporal forecasting for the extended prediction task relative to the baseline model. Fig.~\ref{fig12}.B further demonstrates that the physics-guided network achieves higher fidelity in predicting droplet morphology. These improvements in quantitative metrics indicate that the physics-guided model maintains lower cumulative error over extended iterative rollout. A similar trend is observed for the phase-separation task in Fig.~\ref{fig13}. While both models perform well in iterative rollout of bicontinuous phase dynamics, the physics-guided model consistently achieves superior predictive accuracy.

\begin{figure}[h!]
\centering
\includegraphics[width=0.6\linewidth]{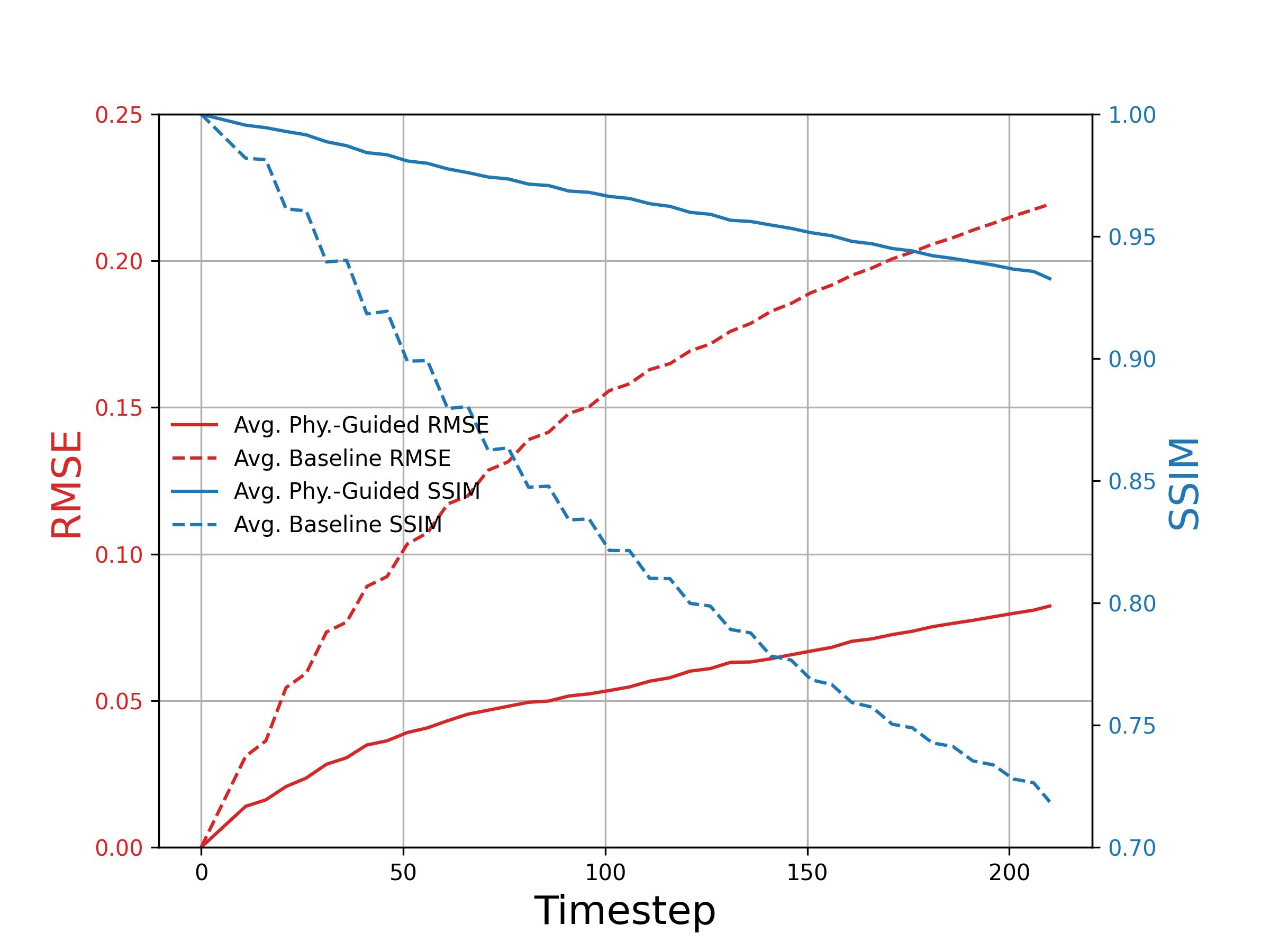}
\caption{\textbf{Accuracy metrics of physics-guided model and baseline model for spinodal decomposition prediction (10 input frames, 200 output frames):} Dataset-averaged RMSE and SSIM evaluated every 5 frames over the prediction horizon. Physics-guided model metrics are represented by the solid line, and the baseline model metrics are represented by the dashed line.}
\label{fig14}
\end{figure}

\begin{figure}[h!]
\centering
\includegraphics[width=\linewidth]{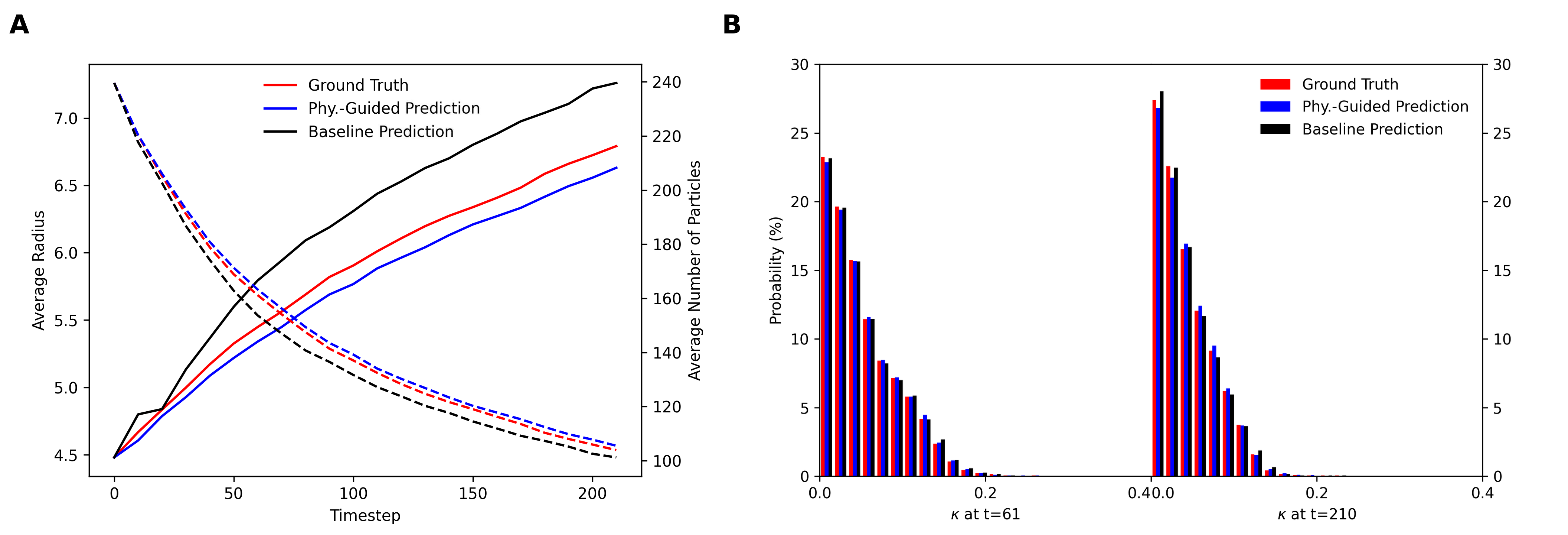}
\caption{\textbf{Physics-based metrics of physics-guided model and baseline model for spinodal decomposition prediction (10 input frames, 200 output frames):} (A) Average radius and average number of particles across $c_{0}=0.25$ samples; (B) Interface segment curvature distribution at timesteps t=61, 210 of samples with concentration $c_{0}=0.5$.}
\label{fig15}
\end{figure}

Dataset-level performance is summarized in Fig.~\ref{fig14}. The mean RMSE and SSIM, computed at 5-frame intervals, indicate consistently strong predictive performance across the dataset. Physics-based metrics are presented in Fig.~\ref{fig15}. In Fig.~\ref{fig15}.A, the average particle radius and particle count are shown for all test samples with concentration $c_{0}=0.25$. At early timesteps, both models closely follow the ground-truth evolution. At later timesteps, however, the physics-guided model remains more consistent with the ground truth than the baseline model. This behavior suggests that physics-guided regularization mitigates error accumulation during iterative rollout, which is a significant challenge for the baseline model. Fig.~\ref{fig15}.B shows the interface curvature distribution for samples with $c_{0}=0.5$ at timesteps $t = 61$ and $t = 210$. Both models closely match the ground-truth distribution, consistent with the baseline model’s strong performance in capturing phase separation dynamics.

\begin{figure}[h!]
\centering
\includegraphics[width=0.7\linewidth]{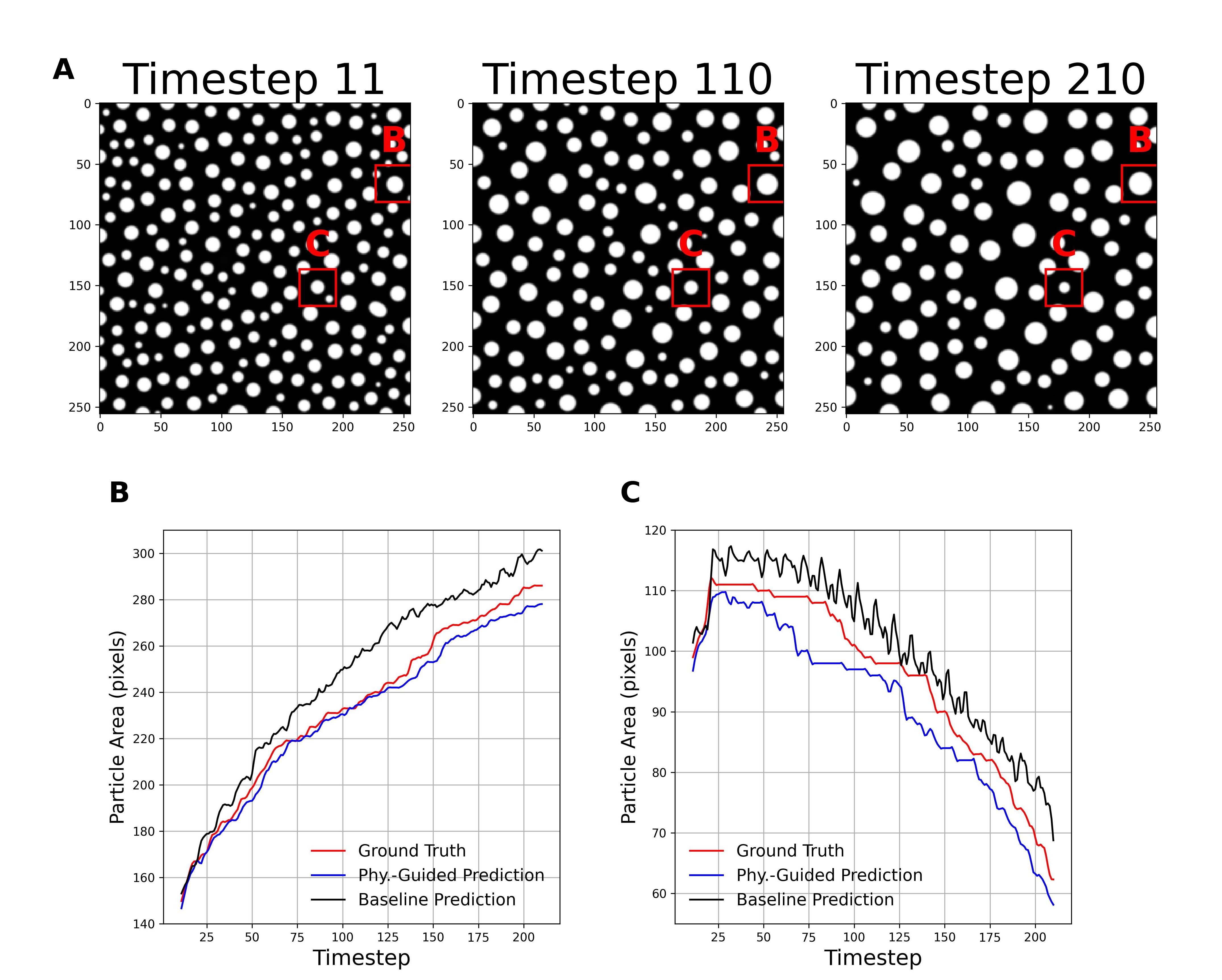}
\caption{\textbf{Tracked particle growth/shrink of physics-guided model and baseline model for spinodal decomposition prediction (10 input frames, 200 output frames):} (A) Tracked particle subjects; (B) growth of tracked particle on ground truth, Physics-guided model, and baseline model; (C) shrinking of tracked particle on ground truth, physics-guided model, and baseline model.}
\label{fig16}
\end{figure}

Fig.~\ref{fig16} compares the temporal evolution of growing and shrinking particles between model predictions and the ground truth sequence. The ground truth frames at $t = 11, 110,$ and $210$ (Fig.~\ref{fig16}.A) are drawn from the same sequence shown in Fig.~\ref{fig12}. Selected particles are highlighted, with the growing and shrinking cases tracked in Fig.~\ref{fig16}.B and Fig.~\ref{fig16}.C, respectively. Particle area is quantified at each timestep and smoothed using a Savitzky–Golay filter for improved interpretability. The physics-guided network more closely follows the ground-truth evolution during both growth and shrinkage, while exhibiting reduced fluctuations in particle size relative to the baseline network. These results highlight the benefit of incorporating physics-based regularization into the model.

\begin{figure}[h!]
\centering
\includegraphics[width=\linewidth]{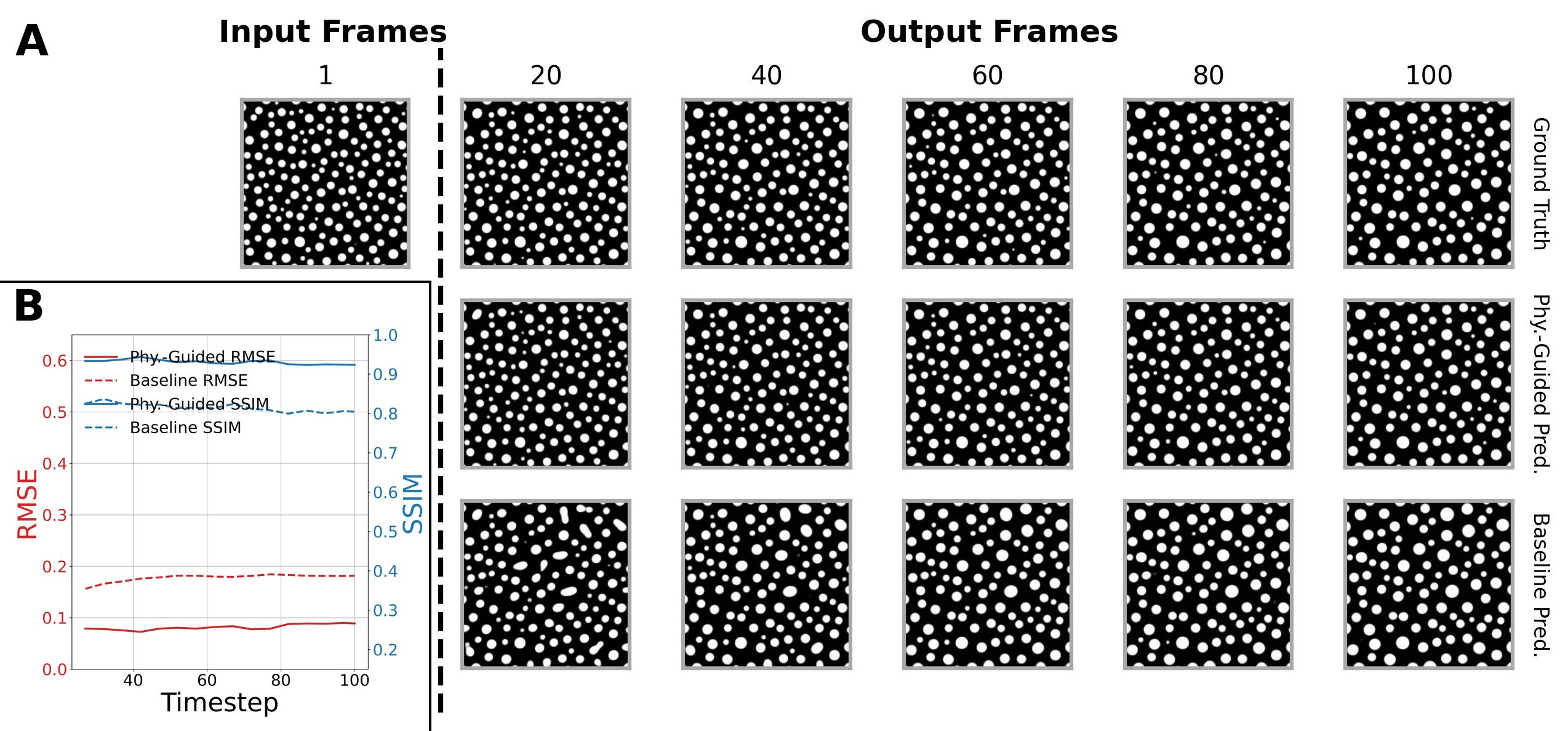}
\caption{\textbf{Spinodal decomposition prediction (1 input frame, 99 output frames) comparison:} Two sets of predictions are displayed along the ground truth. (A) Prediction sequence and the corresponding ground-truth sequence, shown at t = 20, 40, 60, 80, 100; (B) RMSE and SSIM computed between the predicted and ground-truth sequences at 5-frame intervals.}
\label{fig17}
\end{figure}

\begin{figure}[h!]
\centering
\includegraphics[width=\linewidth]{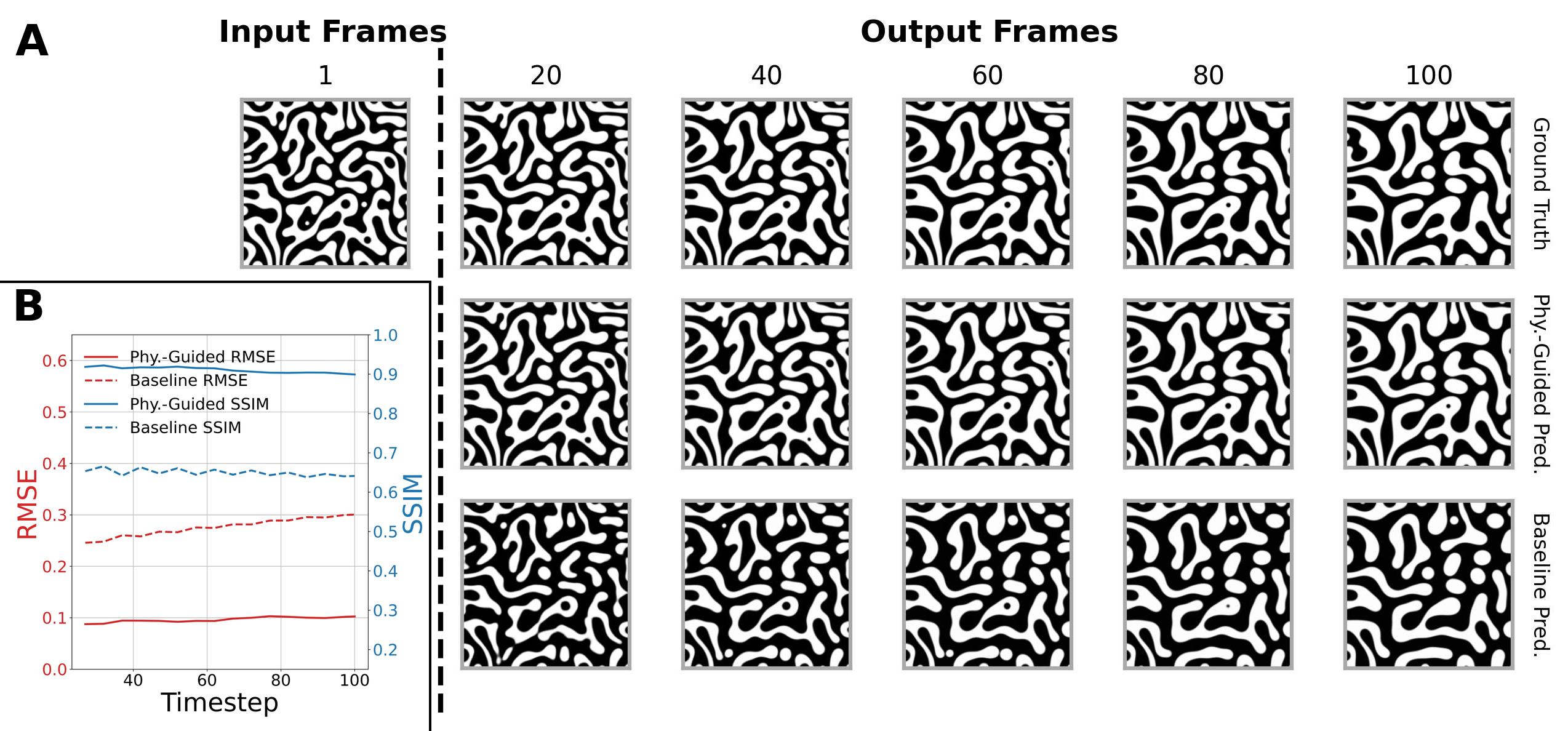}
\caption{\textbf{Spinodal decomposition prediction (1 input frame, 99 output frames) comparison:} Two sets of predictions are displayed along the ground truth. (A) Prediction sequence and the corresponding ground-truth sequence, shown at t = 20, 40, 60, 80, 100; (B) RMSE and SSIM computed between the predicted and ground-truth sequences at 5-frame intervals.}
\label{fig18}
\end{figure}

\subsection{Predictions of 99 future frames from 1 input frame (1$\rightarrow$99)}

For the 1$\rightarrow$99 task, 600 test samples are evaluated on the network models that are trained to predict 90 future frames from 10 input frames. Using iterative roll-out, we can extend the prediction horizon beyond the nominal 90-frame output length. Also, with less temporal context, we apply zero-padding to the 1-frame input, allowing the network to process data that is expected to be 10 frames long. Two representative examples are shown in Figs.~\ref{fig17} ($c_{0}$ = 0.25) and~\ref{fig18} ($c_{0}$ = 0.5). For compact visualization, the input sequence is displayed at $t=1$, while the predicted and ground-truth outputs are shown at $t=20, 40, 60, 80,$ and $100$ (Figs.~\ref{fig17}.A and \ref{fig18}.A). The prediction sequences include the baseline fully convolutional spatiotemporal network, and the physics-guided fully convolutional spatiotemporal network. Under reduced input conditions, the physics-guided network demonstrates a clear performance advantage over the baseline model. The baseline model is particularly sensitive to limited temporal context, resulting in degraded predictive accuracy. In contrast, the physics-guided network maintains robust predictions despite reduced temporal information. This improvement is evident in the prediction sequences shown in Figs.~\ref{fig17}.A and \ref{fig18}.A, and is further supported by quantitative metrics in Figs.~\ref{fig17}.B and \ref{fig18}.B, which indicate higher accuracy for the physics-guided model. These results suggest that physics-based regularization provides inductive bias that compensates for missing temporal information.

\begin{figure}[h!]
\centering
\includegraphics[width=0.5\linewidth]{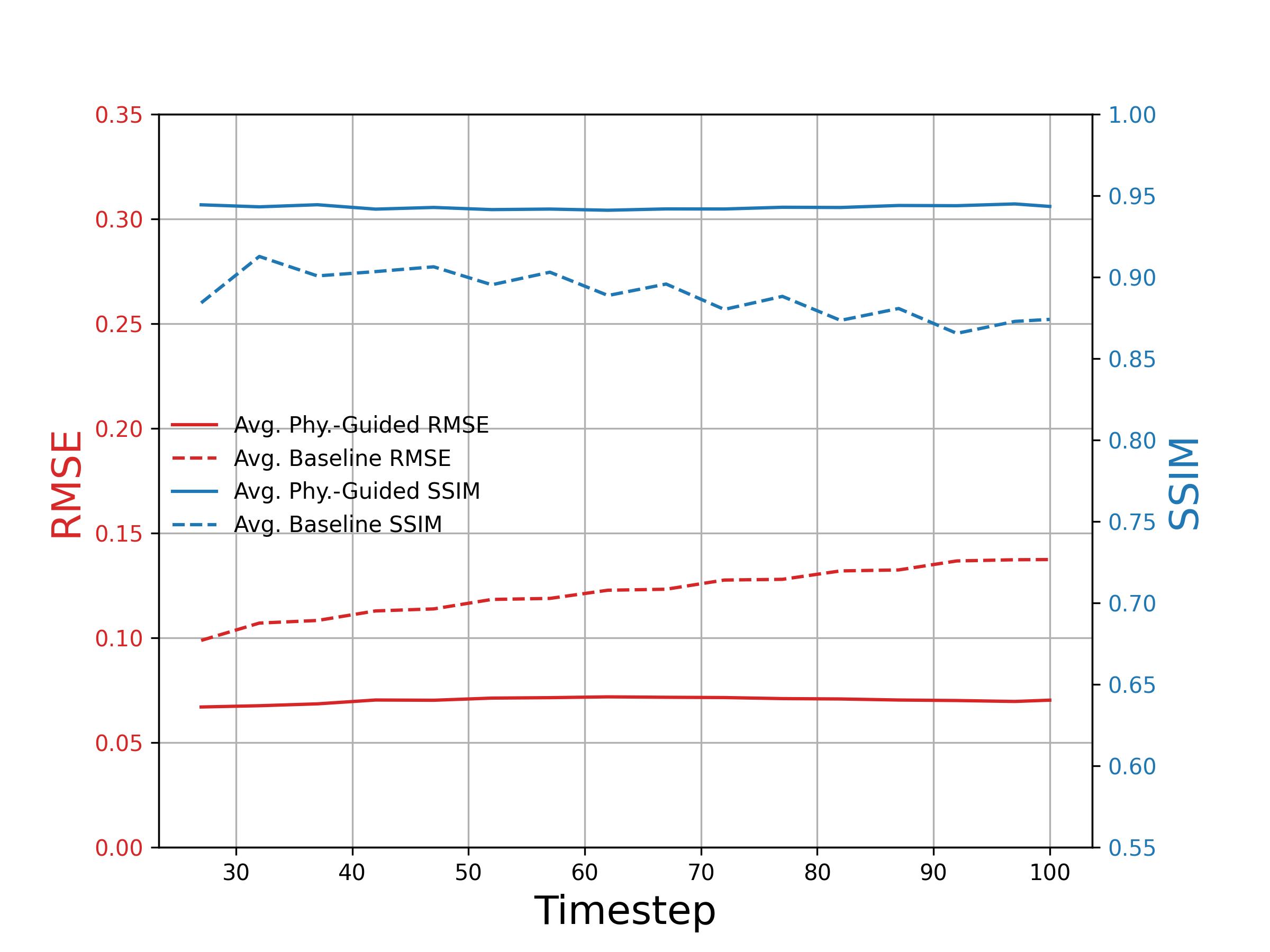}
\caption{\textbf{Accuracy metrics of physics-guided model and baseline model for spinodal decomposition prediction (1 input frame, 99 output frames):} Dataset-averaged RMSE and SSIM evaluated every 5 frames over the prediction horizon. Physics-guided model metrics are represented by the solid line, and the baseline model metrics are represented by the dashed line.}
\label{fig19}
\end{figure}

\begin{figure}[h!]
\centering
\includegraphics[width=\linewidth]{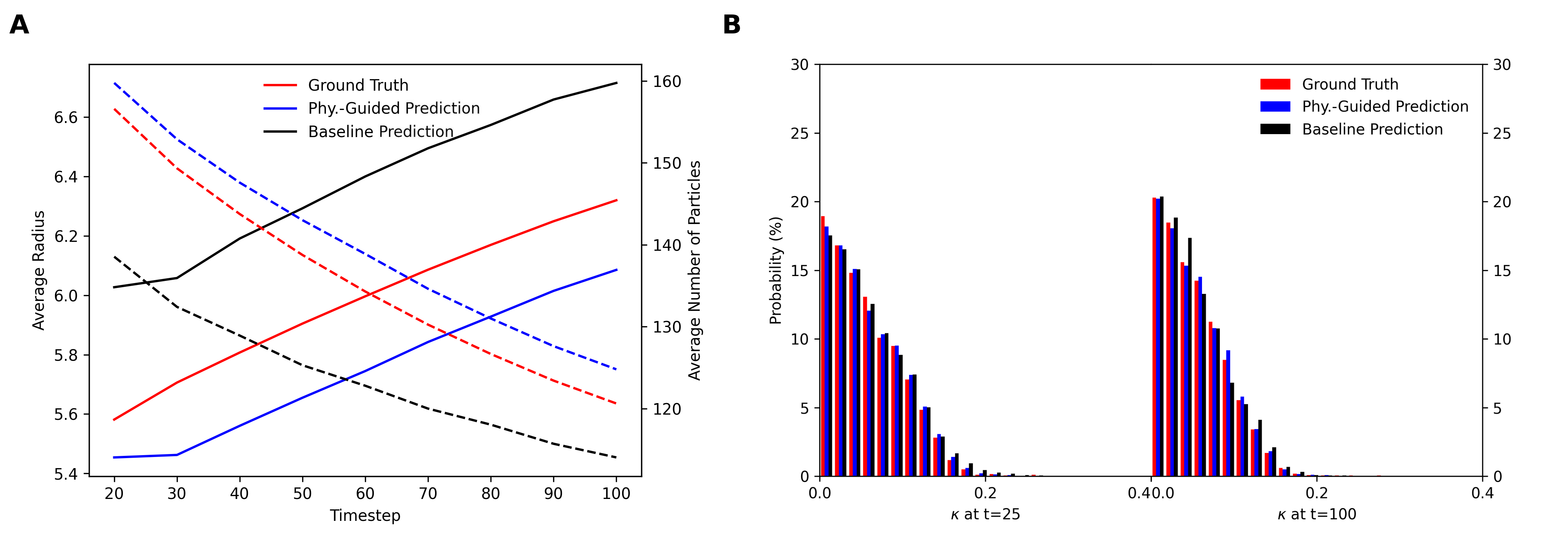}
\caption{\textbf{Physics-based metrics of physics-guided model and baseline model for spinodal decomposition prediction (1 input frame, 99 output frames):} (A) Average radius and average number of particles across $c_{0}=0.25$ samples; (B) Interface segment curvature distribution at timesteps t=25, 100 of samples with concentration $c_{0}=0.5$.}
\label{fig20}
\end{figure}

Dataset-level performance is summarized in Fig.~\ref{fig19}, where mean RMSE and SSIM (computed at 5-frame intervals) indicate strong overall predictive accuracy. Physics-based metrics are presented in Fig.~\ref{fig20}. In Fig.~\ref{fig20}.A, both models exhibit increased deviation in particle radius and count for $c_{0}=0.25$, reflecting the difficulty of prediction under reduced temporal context. However, the physics-guided model remains consistently closer to the ground truth, suggesting improved preservation of key physical statistics. Fig.~\ref{fig20}.B presents the interface curvature distribution for $c_{0}=0.5$ at $t = 25$ and $t = 100$, where both models achieve reasonable agreement with the ground truth distribution.

\begin{figure}[h!]
\centering
\includegraphics[width=0.7\linewidth]{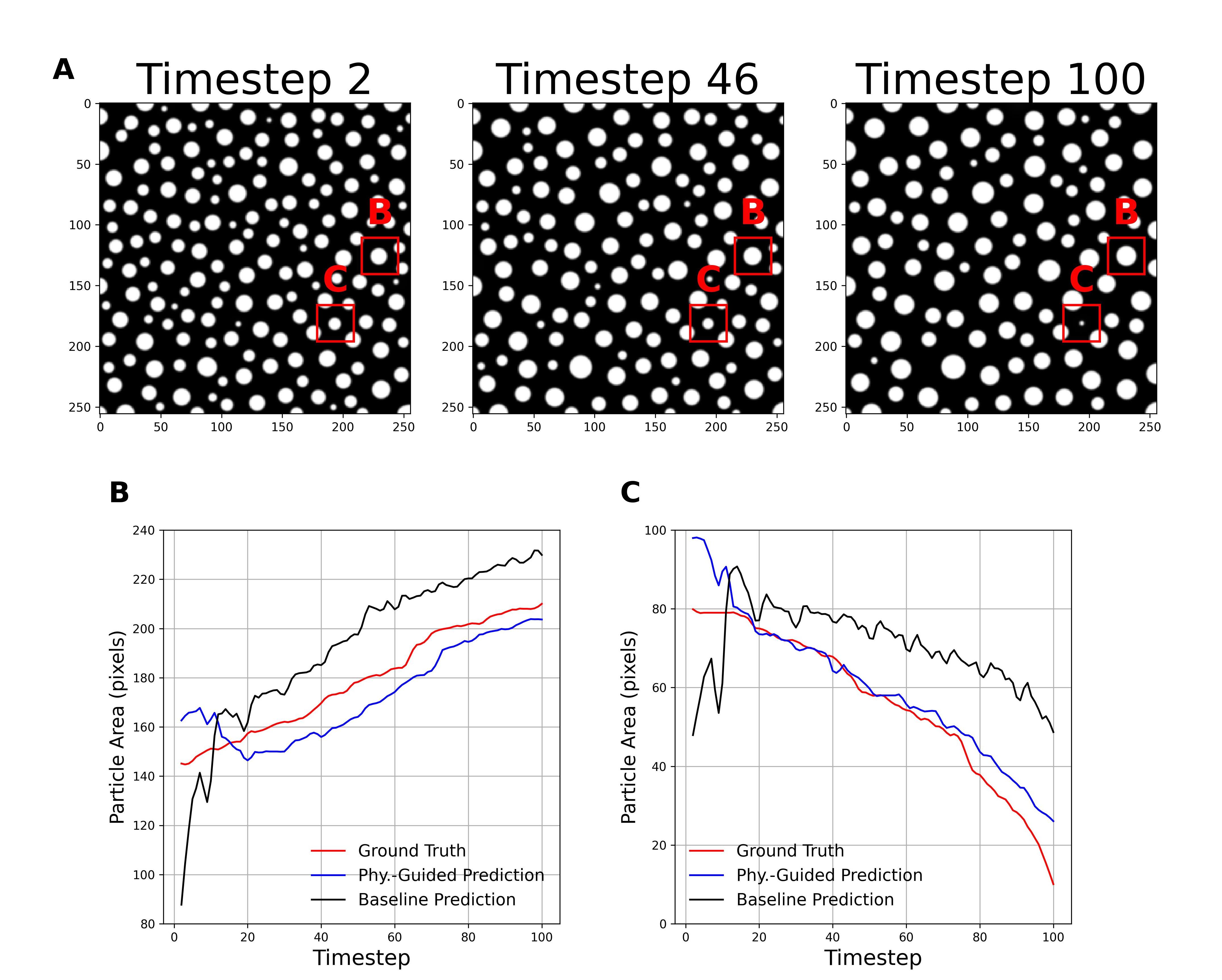}
\caption{\textbf{Tracked particle growth/shrink of physics-guided model and baseline model for spinodal decomposition prediction (1 input frame, 99 output frames):} (A) Tracked particle subjects; (B) growth of tracked particle on ground truth, physics-guided model, and baseline model; (C) shrinking of tracked particle on ground truth, physics-guided model, and baseline model.}
\label{fig21}
\end{figure}

Fig.~\ref{fig21} compares the temporal evolution of growing and shrinking particles between predicted sequences and the ground truth sequence. The ground truth frames at $t = 2, 46,$ and $100$ (Fig.~\ref{fig21}.A) correspond to the same sequence shown in Fig.~\ref{fig17}. Selected particles are highlighted, with the growing and shrinking cases tracked in Fig.~\ref{fig21}.B and Fig.~\ref{fig21}.C, respectively. Particle area is quantified at each timestep and smoothed using a Savitzky–Golay filter for improved interpretability. Despite the reduced temporal context, the physics-guided network more accurately captures particle growth and shrinkage rates than the baseline model. While both models exhibit errors during the initial prediction frames, the physics-guided model shows substantially lower deviation from the ground truth under less temporal context.

\subsection{Computational Efficiency and Inference Speedup}
The physics-based regularizer in the proposed model introduces a modest training overhead relative to SimVPv2, but significantly improves predictive fidelity while preserving a substantially lower training cost than state-of-the-art spatiotemporal models. Table~\ref{tab:runtime} shows that the proposed model is approximately 1.2$\times$ slower than SimVPv2 in training time, yet 4.75$\times$ faster than E3D-LSTM. Inference time remains comparable to SimVPv2 and is 25.8$\times$ faster than E3D-LSTM. Collectively, these results demonstrate that the proposed model achieves an effective balance between accuracy and computational efficiency, making it well-suited for high-throughput, high-resolution spatiotemporal forecasting with iterative rollout.

\begin{table}[h!]
\centering
\begin{tabular}{|c|c|c|}
\hline
\textbf{\makecell{Models \\ (200 Epochs, batch size 1)}} & \textbf{\makecell{Training Time \\ (hours)}} & \textbf{\makecell{Average Inference Time for \\ 100 samples (seconds)}} \\
\hline
The Proposed Model & 97.12 & 2.00 \\
SimVPv2 & 80.75 & 2.00 \\
E3D-LSTM & 461.88 & 51.60 \\
\hline
\end{tabular}
\caption{\textbf{Inference efficiency comparison across deep spatiotemporal models for spinodal decomposition prediction}. Wall-clock training time and inference time for spinodal decomposition prediction. The proposed fully convolutional spatiotemporal model achieves markedly lower inference time than recurrent network (E3D-LSTM).}
\label{tab:runtime}
\end{table}

\section{Discussion and Conclusion}

This work presents a physics-guided fully convolutional spatiotemporal learning framework for microstructure evolution prediction. The proposed approach is motivated by the need to bridge the gap between efficient data-driven forecasting and physically consistent phase-field surrogate modeling. Purely data-driven spatiotemporal models can achieve strong image-level prediction accuracy, but they are not explicitly regularized to remain consistent with the governing physical evolution law. For spinodal decomposition, this limitation is important because the evolution is governed by the Cahn--Hilliard (CH) equation, which encodes mass-conserving, chemical-potential-driven diffusion, interfacial-energy effects, and nonlinear coarsening dynamics. To address this issue, we reformulate the CH equation as a residual constraint and incorporate it into model training as a physics-guided regularization term.

The numerical results demonstrate that the proposed physics-guided model improves prediction accuracy compared with the purely data-driven fully convolutional baseline. The improvement is observed not only in standard image-level metrics such as RMSE and SSIM, but also in physically relevant morphology and statistical quantities, including particle radius, particle count, interface curvature distribution, and tracked particle growth and shrinkage behavior. These results indicate that the benefit of physics guidance extends beyond visual similarity. By penalizing deviations from the governing CH dynamics during training, the model is encouraged to produce microstructure trajectories that better preserve physically meaningful coarsening trends. Importantly, the physics-guided residual is used only during training. At inference time, the model remains a direct fully convolutional predictor and does not require solving the CH equation or performing additional optimization. This preserves the computational efficiency and scalability of fully convolutional spatiotemporal learning. The results also show that the trained model can be directly evaluated on higher-resolution microstructure sequences without architectural modification, highlighting the practical value of the fully convolutional design for spatial-resolution transfer.

The extended-horizon and reduced-context experiments further illustrate the value of the physics-guided formulation. In long-horizon prediction, purely data-driven models may accumulate errors through iterative rollout, leading to gradual deviation from the reference evolution. The proposed physics-guided model reduces this drift and better maintains the morphology and statistics of the evolving microstructure. In reduced-context prediction, where limited temporal information is available, the physics-guided model also exhibits improved robustness. These observations suggest that equation-based residual regularization provides a useful physical inductive bias when the available input information is incomplete or when prediction is required beyond the nominal training horizon.

Despite these advantages, several limitations remain. First, the present study focuses on two-dimensional spinodal decomposition with fixed physical parameters. Although the results demonstrate improved robustness across spatial resolutions and temporal prediction settings, the current experiments do not establish generalization across different material parameters, free-energy models, or mobility functions. Extending the framework to parameterized phase-field systems will be important for broader materials design applications. Second, the CH residual is evaluated using discrete approximations on predicted image sequences. The accuracy and stability of the residual may depend on the choice of spatial discretization, temporal sampling, boundary treatment, and the relative weight assigned to the physics-guided loss. Future work should systematically investigate these numerical and optimization factors. Third, the current framework is evaluated for spinodal decomposition only. Extension to other microstructure evolution processes, such as grain growth, precipitation, solidification, and multiphase systems, would further test the generality of the approach. 
Another important direction is uncertainty quantification and data assimilation. For digital-twin-enabled materials modeling, a surrogate model should not only forecast future microstructure states efficiently, but also quantify prediction uncertainty, assimilate new observations, and adapt to changes in processing conditions or material parameters. The proposed physics-guided framework provides a foundation for this goal because it combines fast neural-network inference with equation-based physical guidance. Future extensions could incorporate Bayesian or ensemble-based uncertainty estimation, parameter-conditioned learning, adaptive residual weighting, and online correction using newly observed microstructure data.

In conclusion, this study develops a physics-guided fully convolutional spatiotemporal learning framework for accelerated microstructure evolution prediction. By incorporating the CH residual into the training objective, the proposed model improves predictive fidelity, reduces long-horizon drift, and better preserves physically meaningful morphology and coarsening statistics compared with a purely data-driven baseline. At the same time, the method retains the fast inference and spatial scalability of fully convolutional architectures because no phase-field solver is required during prediction. These results demonstrate the potential of physics-guided spatiotemporal learning as an efficient and physically informed surrogate modeling strategy for microstructure evolution, and they provide a step toward digital-twin-enabled materials modeling and design.

\section*{Declaration of competing interest} 
The authors declare that they have no known competing financial interests or personal relationships that could have appeared to influence the work reported in this paper.

\section*{Code and Data Availability}
The code developed for this work and the datasets generated and analyzed during the current study can be found here: \url{https://github.com/mtrimboli2018/CNN_STL_MicEvo}.

\section*{Declaration of generative AI and AI-assisted technologies in the manuscript preparation process} 
During the preparation of this work the author(s) used ChatGPT in order to improve language and readability. After using this tool/service, the author(s) reviewed and edited the content as needed and take(s) full responsibility for the content of the publication.

\section*{Acknowledgment}
X. Li's work was partially funded by the Division of Mathematical Sciences, National Science Foundation with the award numbers 2410678.

\bibliographystyle{elsarticle-num} 

\bibliography{reference}

\end{document}